\documentclass[lettersize,journal]{IEEEtran}
\usepackage{amsmath,amsfonts}
\usepackage{algorithmic}
\usepackage{array}
\usepackage[caption=false,font=normalsize,labelfont=sf,textfont=sf]{subfig}
\usepackage{textcomp}
\usepackage{stfloats}
\usepackage{url}
\usepackage{verbatim}
\usepackage{graphicx}
\def\BibTeX{{\rm B\kern-.05em{\sc i\kern-.025em b}\kern-.08em
    T\kern-.1667em\lower.7ex\hbox{E}\kern-.125emX}}
\usepackage{balance}

\newcolumntype{M}[1]{>{\centering\arraybackslash}m{#1}}
\usepackage[numbers]{natbib} %推荐使用
\PassOptionsToPackage{colorlinks,bookmarksopen,bookmarksnumbered,citecolor=green,linkcolor=red,urlcolor=green}{hyperref}
\usepackage{amsmath} %公式粗体包
\usepackage{amsfonts,amssymb} %空心字母包
\usepackage{makecell} % 用于多行标题
\usepackage{array} %旋转表格
\usepackage{bbding} % √ 与 叉 “×”
\usepackage{wasysym}
\usepackage{bbm}
\usepackage{dsfont}
\usepackage{bbold}
\usepackage{lipsum}
\usepackage{booktabs}
\usepackage{mathtools}
\usepackage{balance}
\usepackage{color}
\usepackage[ruled,vlined,linesnumbered]{algorithm2e}
\usepackage{diagbox}
\usepackage{epstopdf}
\usepackage{pifont}
\usepackage{fixltx2e}
\usepackage{multirow}
\usepackage{url}
\usepackage{verbatim}
\usepackage{footmisc}
\usepackage{physics}
\usepackage[skip=3pt, font={small}]{caption} % example skip set to 2pt
\usepackage{tabularx}
\usepackage{tikz}
\usepackage{amsthm}
\usepackage{wrapfig}
\usepackage{tabulary}
\usepackage{booktabs}
\usepackage{bbm}
\usepackage{enumitem} % Make sure this line is present if you're using enumitem features
\usepackage{threeparttable}
\usepackage[T1]{fontenc}
\usepackage[utf8]{inputenc}
\usepackage{hyperref}

\usepackage{tikz,xcolor,hyperref}% Make Orcid icon
\definecolor{lime}{HTML}{A6CE39}
\DeclareRobustCommand{\orcidicon}{%
	\begin{tikzpicture}
		\draw[lime, fill=lime] (0,0) 
		circle [radius=0.16] 
		node[white] {{\fontfamily{qag}\selectfont \tiny ID}};    \draw[white, fill=white] (-0.0625,0.095) 
		circle [radius=0.007];    \end{tikzpicture}
	\hspace{-2mm}}
\foreach \x in {A, ..., Z}{%
	\expandafter\xdef\csname orcid\x\endcsname{\noexpand\href{https://orcid.org/\csname orcidauthor\x\endcsname}{\noexpand\orcidicon}}
}

\begin{document}
\title{Adaptive-LIO: Enhancing Robustness and Precision through Environmental Adaptation in LiDAR Inertial Odometry}

\author{IEEE Publication Technology Department
	\thanks{Manuscript created October, 2020; This work was developed by the IEEE Publication Technology Department. This work is distributed under the \LaTeX \ Project Public License (LPPL) ( http://www.latex-project.org/ ) version 1.3. A copy of the LPPL, version 1.3, is included in the base \LaTeX \ documentation of all distributions of \LaTeX \ released 2003/12/01 or later. The opinions expressed here are entirely that of the author. No warranty is expressed or implied. User assumes all risk.}}

%\author{Vlad~Niculescu,
%	Tommaso~Polonelli,    ~\IEEEmembership{Member,~IEEE,} Michele~Magno,~\IEEEmembership{Senior Member,~IEEE,}
%	and~Luca~Benini,~\IEEEmembership{Fellow,~IEEE}%
%	
%	\thanks{V. Niculescu and L. Benini are with the Integrated Systems Laboratory of ETH Z\"urich, ETZ, Gloriastrasse 35, 8092 Z\"urich, Switzerland (e-mail: vladn@iis.ee.ethz.ch, lbenini@iis.ee.ethz.ch).}%
%	
%	\thanks{T. Polonelli and M. Magno are with the Center for Project-Based Learning of ETH Z\"urich, ETZ, Gloriastrasse 35, 8092 Z\"urich, Switzerland (e-mail: tommaso.polonelli@pbl.ee.ethz.ch, michele.magno@pbl.ee.ethz.ch).}%
%	
%	\thanks{L. Benini is also with the Department of Electrical, Electronic and Information Engineering of University of Bologna, Viale del Risorgimento 2, 40136 Bologna, Italy.}}\\

\author{Chengwei Zhao$^{*}$\orcidA{},
	Kun Hu$^{*}$\orcidB{}, 
	Jie Xu$^{\dag}$\orcidC{}, 
	Lijun Zhao$^{\dag}$\orcidD{},\\
	Baiwen Han\orcidE{}, 
		Kaidi Wu\orcidG{},
	Maoshan Tian\orcidF{}, 
	Shenghai Yuan\orcidH{}

	% \thanks{This work was supported in part by the National Natural Science Foundation (62073101) of China and Self-Planned Task (SKLRS202301A09, SKLRS202417B) of SKLRS(HIT) of China.}
	\thanks{This work was supported by State Key Laboratory of Robotics and Systems (HIT) and  the National Natural Science Foundation of China under Grand 62073101.}

	\thanks{Chengwei Zhao, Lijun Zhao and Baiwen Han are with the State Key Laboratory of Robotics and Systems, Harbin Institute of Technology, Harbin, 150001, HeiLongJiang, China (e-mail:zhaolj@hit.edu.cn, 23s136401@stu.hit.edu.cn).}%

	\thanks{Chengwei Zhao is with Hangzhou Qisheng Intelligent Techology Co. Ltd., 4083 Jianshe Fourth Road, Hangzhou, 311217, Zhejiang, China (e-mail: zhaochengwei@qishengrobot.com).}%

	\thanks{Kun Hu and Kaidi Wu are with the School of Mechatronic Engineering, China University of Mining and Technology, XuZhou, 221116, JiangSu, China (e-mail: ts22050017a31@cumt.edu.cn, ts23050205p31@cumt.edu.cn).}%

	\thanks{Maoshan Tian is with the School of Mechanical and Electrical Engineering, University of Electronic Science And Technology, Chengdu, 611731, Sichuan, China (e-mail: 202222040706@std.uestc.edu.cn).}%

	\thanks{Jie Xu and Shenghai Yuan are with the School of Electrical and Electronic Engineering, Nanyang Technological University, 50 Nanyang Avenue, 639798, Singapore (e-mail: jie.xu@ntu.edu.sg; shyuan@ntu.edu.sg).}

	\thanks{$^{*}$Chengwei Zhao and Kun Hu contributed equally to this work and should be considered co-first authors.}

	\thanks{$^{\dag}$Jie Xu and Lijun Zhao are the common corresponding authors.}
}

\markboth{Journal of \LaTeX\ Class Files,~Vol.~18, No.~9, September~2020}%
{How to Use the IEEEtran \LaTeX \ Templates}

\maketitle

\begin{abstract}
	The emerging Internet of Things (IoT) applications, such as driverless cars, have a growing demand for high-precision positioning and navigation. Nowadays, LiDAR inertial odometry becomes increasingly prevalent in robotics and autonomous driving. However, many current SLAM systems lack sufficient adaptability to various scenarios. Challenges include decreased point cloud accuracy with longer frame intervals under the constant velocity assumption, coupling of erroneous IMU information when IMU saturation occurs, and decreased localization accuracy due to the use of fixed-resolution maps during indoor-outdoor scene transitions. To address these issues, we propose a loosely coupled adaptive LiDAR-Inertial-Odometry named \textbf{Adaptive-LIO}, which incorporates adaptive segmentation to enhance mapping accuracy, adapts motion modality through IMU saturation and fault detection, and adjusts map resolution adaptively using multi-resolution voxel maps based on the distance from the LiDAR center. Our proposed method has been tested in various challenging scenarios, demonstrating the effectiveness of the improvements we introduce. The code is open-source on GitHub: \href{https://github.com/chengwei0427/adaptive_lio}{Adaptive-LIO}.

\end{abstract}


\section{Introduction}\label{chap_1_intro}
\IEEEPARstart{A}{ccurate} and continuous navigation solutions have become critical in some location-driven Internet of Things (IoT) applications, such as autonomous driving. However, it is often difficult for a standalone sensor to meet the needs of a high-precision navigation in complex scenarios. Thus, multisensor fusion, which can take advantage of complementary properties from heterogeneous sensors, may be a feasible solution to ensure navigation performance in urban environments. Localization provides state feedback to the robot's onboard controller, while dense 3D maps provide essential information for trajectory planning, such as free space and obstacles.

Currently, many works employ tightly-coupled systems, such as LIO-SAM \cite{shan2020lio}, LINS \cite{qin2020lins}, FAST-LIO2 \cite{xu2022fast}. However, in some cases, it is challenging to achieve IMU and LiDAR extrinsic calibration and hardware time synchronization. For instance, defects in the design of LiDAR or IMU devices may prevent accurate extrinsic calibration data or hardware timestamps. Additionally, time delays and jitter in high-dynamic scenes can degrade the accuracy of hardware time synchronization, while environmental instability complicates hardware time synchronization. Therefore, tightly-coupled systems may not perform well on such hardware devices and may not adapt to various types of scenes. Consequently, efficient, accurate, and adaptive LiDAR-based localization and mapping remain challenging, as detailed below.

\begin{itemize}
	\item \textbf{Challenge 1:} Traditional Continuous Time (CT) models typically assume constant velocity over longer periods, which can lead to significant state errors and affect the accuracy of the LiDAR point cloud.

	\item \textbf{Challenge 2:} The IMU may sometimes exceed its measurement range or malfunction, causing tightly coupled SLAM systems to fail and compromising system stability.

	\item \textbf{Challenge 3:} The point cloud density differs between open and narrow areas. Using the same resolution for regions of varying sizes can degrade map accuracy, which can easily result in degeneration.
\end{itemize}

In this work, we address these issues through 3 novel framework: adaptive frame segmentation, IMU saturation and fault detection, and multi-resolution maps. More specifically, our contributions are as follows:
\begin{itemize}
	\item \textbf{Adaptive Frame Segmentation:} The accuracy of LiDAR odometry is determined by the degree of motion distortion. Shorter frame intervals approach actual motion conditions more closely, resulting in more accurate pose estimation. We apply adaptive frame segmentation to determine whether frame segmentation is necessary and  how many frames to segment, thereby reducing point cloud distortion and improving system accuracy and robustness.
	\item \textbf{Adaptive Motion Mode Switching:} We adopt a loosely coupled model to ensure system robustness. When IMU issues are detected, the system switches to LO modality, and when the IMU is operating normally, it switches back to LIO modality.
	\item \textbf{Adaptive Multi-Resolution Maps:} We employ a multi-resolution voxel map to ensure consistency in map density. This involves using low-resolution maps in open areas and high-resolution maps in narrow areas. This ensures consistent overall map sparsity across scenes at different distances from the LiDAR, enhancing mapping accuracy and reducing degeneration occurrences.
	\item We integrate these three adaptive techniques into the proposed loosely coupled LiDAR inertial odometry system Adaptive-LIO and open source it on Github for the benefit of the community. Finally we conduct experiments on open source datasets and real scenarios to verify the robustness and accuracy of the system in challenging scenarios.
\end{itemize}

The remainder of the paper is organized as follows: in chapter \ref{chap_2_related_works}, we discuss related works. We outline the complete system flow and details of each key component in chapters \ref{chap_3_priliminary}, \ref{chap_4_system_overview}, \ref{chap_5_adaptive_frame_segmentation}, \ref{chap_6_adaptive_motion_switching}, \ref{chap_7_adaptive_multiresolution_map}. Experiments are presented in \ref{chap_8_experiment}. Chapter \ref{chap_9_conclusion} is the conclusion.

\section{Related Works}\label{chap_2_related_works}

\subsection{LiDAR inertial odometry}
Accurately and efficiently estimating odometry information in GPS-constrained, unknown underground environments for robot motion planning and control remains challenging in the field of robotics \cite{zhao2023autonomous}. State estimation strategies for these environments \cite{neumann2014towards, papachristos2019autonomous} typically rely on proprioceptive sensing (e.g., IMUs, wheel encoders) and active exteroceptive methods (e.g., GNSS \cite{chen2023enhancing}, Radar \cite{du2024general}, LiDAR, Time-of-Flight cameras, UWB \cite{li2023multimodal, papadimitriou2022range, pan2022indoor}).

Currently, most LiDAR SLAM algorithms can be categorized into two main approaches: indirect methods and direct methods, based on whether feature extraction from LiDAR frame is required. In the \textbf{indirect methods}, Zhang et al proposed LOAM \cite{zhang2014loam}, a real-time method for odometry and mapping that extracts edge and planar features; the separation structure between frame-to-frame and frame-to-map in LOAM has been adopted by many subsequent works such as Lego-LOAM \cite{shan2018lego}, which accounts for the constraints induced by the ground during frame-to-frame matching to improve the accuracy of the odometry; LIO-SAM \cite{shan2020lio}, which first formulated the LIO odometry as a factor graph, a form that allows relative and absolute measurements from different sources, including closed-loop measurements, to be factored into the system. 3D-CSTM \cite{cong20223d} proposes an automatic structural feature extraction method based on scanline characteristics and a uniform observation selection method based on Directional Constraint Sphere. To address the challenges of enabling SLAM algorithms on resource-constrained processors, NanoSLAM \cite{niculescu2023nanoslam} proposes a lightweight and optimized end-to-end SLAM method, specifically designed to run on centimeter-scale robots with a power consumption of only 87.9mW. The algorithm, designed to leverage the parallel capabilities of the RISC-V processing cores, enables mapping of a general environment with an accuracy of 4.5 cm and an end-to-end execution time of less than 250 ms; FAST-LIO employs the technique of solving Kalman gain \cite{sorenson1966kalman} to avoid the computation of higher-order inverse matrix, which greatly reduces the computational burden; SemanticCSLAM \cite{li2024semanticcslam} proposes a collaborative SLAM system based on environmental semantic landmarks for position recognition and loop closure detection. 
The proposed system consists of a trajectory alignment module and a global optimization module based on environmental landmarks. Through the inertial measurement unit (IMU) carried by the agent, such as an unmanned ground vehicle (UGV), the environment landmarks can be detected. Based on these environmental landmarks, the alignment module aligns trajectories from different agents. To address the issue where the state at the end of the previous frame cannot consistently satisfy constraints from both LiDAR and IMU, leading to inaccuracies in the next frame and local inconsistencies in the estimated state, CT-ICP \cite{dellenbach2022ct} adds logical constraints to make the start state close to the end state of the previous frame, and this inter-frame constraint allows for pose estimation and aberration alignment to be performed at the same time; Kiss-ICP \cite{vizzo2023kiss} improves efficiency and accuracy based on CT-ICP; although the inter-frame alignment method can reduce the computational load of LiDAR odometry, it accumulates a larger cumulative error; furthermore, inter-frame alignment requires a large overlap region between consecutive frames, which is not applicable to solid-state LiDARs with small fields of view. To address these issues, \textbf{direct methods} have been widely adopted, based on FAST-LIO, FAST-LIO2 \cite{xu2022fast} builds on an efficient tightly-coupled iterative Kalman filter to directly register raw points to the map without extracting features, thereby improving mapping and localization accuracy. DLIO \cite{chen2023direct} employs a new coarse-to-fine method to build continuous time trajectories that enables accurate motion correction. This results in fast and parallelizable point-by-point deskewing. Additionally, by performing motion correction and prior generation simultaneously, and by directly registering each frame to the map while bypassing the frame-to-frame step, mapping efficiency is significantly improved; IG-LIO \cite{chen2024ig} proposes a tightly-coupled LiDAR-Inertial-Odometry (LIO) algorithm based on incremental GICP, which integrates the GICP constraints and inertial constraints into a unified estimation framework, IG-LIO uses a voxel-based surface covariance estimator to estimate the surface covariance of the frames and utilizes incremental voxel maps to represent the probabilistic model of the surroundings, and these methods successfully reduce the time consumption of covariance estimation, nearest neighbor search, and map management. Li et al. \cite{li2022tightly}  proposes a tightly coupled multi-GNSS precise point positioning (PPP)/INS/LiDAR system that utilizes a LiDAR sliding-window plane-feature tracking method to further enhance navigation accuracy and computational efficiency. The proposed GNSS/INS/LiDAR tightly coupled system can maintain sub-meter level horizontal positioning accuracy in GNSS-challenging environments.

However, regardless of whether direct or indirect methods are used, LiDAR-based methods produce poor estimates in scenarios that are not sufficiently geometrically-rich to constrain the motion estimation \cite{gelfand2003geometrically, zhang2016degeneracy, censi2007accurate}. This is particularly evident in tunnel-like environments \cite{neumann2014towards, papachristos2019autonomous}, where it is difficult to constrain relative motion along the main shaft of the tunnel. Techniques used to mitigate these problems include observability-aware point cloud sampling \cite{zhang2017enabling}, degeneration detection and mitigation \cite{zhang2016degeneracy, zhang2017enabling, hinduja2019degeneracy}, and fusion of other information sources from the IMU. This work proposes Adaptive-LIO, an adaptive LiDAR inertial observability perception algorithm suitable for localization and mapping in GPS-constrained, unknown underground environments. To address potential observability issues, we use geometric observability scoring \cite{gelfand2003geometrically} and frame segmentation \cite{yuan2022sr} techniques that allow Adaptive-LIO to be able to observe the scene based on the geometry of its observed predict potential degeneration of its output. The geometric observability score allows us to determine the degree of observability of the scene, which in turn allows the system to determine whether to segment frames and how many frames to segment, ultimately improving the accuracy of mapping.

\subsection{Frame segmentation}
The method of frame segmentation utilizes the continuous scanning characteristic of rotating LiDAR to segment and reconstruct the original input frames, thereby obtaining reconstructed  frames at a higher frequency. The increase in frequency shortens the time interval between consecutive  frames, reducing the time interval for IMU measurement integration and increasing the frequency of state updates. Therefore,  frame reconstruction not only improves the frequency of output poses but also enhances the accuracy of state estimation in the LIO system. To fully account for the highly dynamic motion patterns of portable devices and changes in the structure of the scanning scene, AFLI-Calibr \cite{wu2023afli} employs a LiDAR odometry with adaptive frame length to perform LiDAR-IMU extrinsic calibration. Unlike LiDAR odometry methods with fixed frame lengths, the LiDAR frame length is dynamically adjusted based on the sensor's motion state and the stability of scene matching. SDV-LOAM \cite{yuan2023sdv} and SR-LIVO \cite{yuan2024sr} adopt the frame segmentation reconstruction technique to align the reconstructed frames with the timestamps of the images. This enables the LIO module to accurately determine the state at every imaging moment, thereby improving pose accuracy and processing efficiency. SR-LIO \cite{yuan2022sr}, based on the iterative extended Kalman filter (iEKF) framework, uses a fixed frame segmentation and reconstruction method to segment and reconstruct the original input frames from the rotating LiDAR to obtain higher-frequency reconstructed  frames. Traj-LO \cite{zheng2024traj} combines the geometric information of LiDAR points with the kinematic constraints of trajectory smoothness to restore the spatiotemporal consistent motion of LiDAR.

\subsection{State estimation under IMU saturation}
The fusion of IMU measurements with LiDAR point alignment is usually done using two dominant methods, namely loose coupling and tight coupling. Tightly coupled methods usually have higher robustness and accuracy than loosely coupled methods. However, in all of the above tightly-coupled methods, the IMU data is used as an input to the kinematic model, and thus can be propagated or pre-integrated in EKF propagation for a frame duration, and such EKF propagation or pre-integration will encounter saturation problems if the robot motion is beyond the range of the IMU measurements.

Point-LIO \cite{he2023point} was the first to develop a tightly coupled EKF framework capable of handling saturated IMU measurements in extremely intense motions, such as vibrations and high-speed motions, by modeling IMU measurements with a stochastic process model. The model is extended to system kinematics and IMU measurements are considered as system outputs. Even if the IMU is saturated, the stochastic process-augmented kinematic model allows for smooth estimation of the system state, including angular velocity and linear acceleration; in \cite{quenzel2021real, le2019in2lama, le2020in2laama}, the IMU data is used to provide measurements of angular velocity and linear acceleration predicted from a continuous-time trajectory, based on which the trajectory parameters are optimized along with the LiDAR scanning alignment factors. Treating IMU data as measurements of model outputs can naturally address IMU saturation caused by intense motion, although this ability is limited by the continuous-time model described above; unlike Point-LIO, our approach handles extreme by setting up an IMU saturation detection mechanism that switches the state of the system by determining whether the angular velocity and linear acceleration of the IMU are out of range. Saturated IMU measurements in intense motions, such as vibration and high-speed motions, if the IMU is out of range, the system switches from the LIO modality to the LO modality, otherwise it will keep the LIO modality to continue the pose estimation. To the best of our knowledge, the method for switching between LO and LIO modality has not been proposed in any previous work.

\subsection{Voxel map}
The map structure of LiDAR SLAM is typically broadly categorized into tree-based and voxel-based structures. FAST-LIO2 \cite{xu2022fast} proposes an incremental k-d tree, ikd-Tree, as a map structure. Due to this efficient incremental mapping structure, the FAST-LIO2 \cite{xu2022fast} is able to perform odometry and map construction in real-time at a rate of 10 Hz on rotating LiDARs and 100 Hz on solid-state LiDARs, even on low-power ARM-based computers; inspired by the ikdtree, Faster-LIO \cite{bai2022faster} uses a sparse voxel-based near-neighbor structure iVox (incremental voxels), which can effectively reduce the time-consuming point cloud registration while ensuring the accuracy of LIO; Voxelmap \cite{yuan2022efficient} proposes a highly efficient probabilistic adaptive voxel construction method, which proposes a map composed of voxels, and each voxel contains a planar (or edge) feature capable of realizing the probabilistic representation of the environment and the new voxels. Probabilistic representation of the environment and accurate alignment of new LiDAR frames, and then uses hash-table based voxel maps and octrees to efficiently construct and update the maps. LiTAMIN and LiTAMIN2 \cite{yokozuka2020litamin, yokozuka2021litamin2} proposed a fast registration method by reducing the number of registration points and introducing symmetric KL scattering into the conventional ICP. They were inspired by the NDT method, which first divides the points into separate voxels and then performs the NDT transform in each voxel. To address the issue of precise localization in dynamic environments, the Long-term Static Mapping and Cloning
Localization (LSMCL) \cite{lee2024lsmcl} utilizes 3D LiDAR to provide stable global localization in the presence of dynamic changes in the navigation space. The Long-term Static Mapping (LSM) \cite{10202252} creates a map of static natural landmarks from the measurement data obtained by the LiDAR  in the surrounding environment. Additionally, the proposed Clone Localization (CL), using the map created by LSM, enables accurate localization despite dynamic changes. AdaLIO \cite{10202252} adopts an adaptive map parameter setting strategy that switches between different map parameters based on the number of sampled points in the point cloud and the distance from the occupied grid to the origin.

In contrast to AdaLIO, in our work, we propose a hash table-based adaptive multi-resolution voxel map. This map adaptively allocates map resolutions based on the distance of map points from the LiDAR coordinate system, allowing multiple resolution point cloud to simultaneously participate in building constraints, which in turn ensures the consistency of the density of the map.

\section{Priliminary}\label{chap_3_priliminary}
\subsection{Notation}
\newlength{\tblwidth}
\setlength{\tblwidth}{0.9\linewidth}
%\begin{table}[width=.9\linewidth,cols=4,pos=h]
\begin{table}[h]
	\caption{\textsc{Definitions of Important Variables.}}\label{tbl1}
	\begin{tabular*}{\tblwidth}{ll}
		\toprule
		Symbols       &    Meaning\\
		\midrule
		$t_{b}^k,t_{e}^k$ & First and last timestamp of the $k$-th frame. \\
		$\mathbf{X},\hat{\mathbf{X}}$ & Ground-true, propagated value of state $\mathbf{X}$. \\
		$\mathbf{X}_b^k,\mathbf{X}_e^k$ & State in head and tail of the $k$-th frame.\\
		$\check{\mathbf{X}},\mathbf{\overline{X}}$ &  Optimized and updated value of state \\
		$\delta{\mathbf{X}}$ & Error value of state $\mathbf{X}$.\\
		\bottomrule
	\end{tabular*}
\end{table}
We denote $(.)^W$, $(.)^L$ and $(.)^B$ as 3D points represented in the world coordinate system, LiDAR coordinate system, and IMU coordinate system, respectively. The world coordinates coincide with the origin of the $(.)^B$ frame. The operators $\boxplus$ and $\boxminus$ denote the generalized addition and subtraction defined on manifolds \cite{xu2022fast}.

We define a point cloud frame $\mathbf{F}_k$  composed of the point cloud  $\mathcal{P}_k$, the state $\mathbf{X}=\left\{\mathbf{X}_b^k,\mathbf{X}_e^k\right\}$, the start time $t_b^k$, and the end time $t_e^k$. $\mathbf{X}_b^k$ represents the initial state of the frame, and $\mathbf{X}_e^k$ represents the final state. $\mathcal{P}_k$ is a single-frame point cloud obtained from the time $t_b^k$  to the time $t_e^k$, marked with index $k$. The point cloud $\mathcal{P}_k$ consists of points $\mathbf{p}_k^n\in R^{3\ }$ measured at times $\mathrm{\Delta}t_{k\ }^n$ relative to the start of the frame, where $n=1,...,N$, and $N$ is the total number of points in the frame. The state of the robot in frame $k$ is defined as
\begin{equation}
	\begin{aligned} \mathbf{X}_{(.)}^k=\left[^W\mathbf{q}_{I},^W\mathbf{p}_{I},^W\mathbf{v}_{I},\mathbf{b}_{g},\mathbf{b}_{a}\right]^{\top},
		\label{eq_1_state}
	\end{aligned}
\end{equation}
where $\mathbf{X}_{\left(.\right)}^k$  is $\mathbf{X}_b^k$ or $\mathbf{X}_e^k$, and $^W\mathbf{p}_{I} \in \mathbb{R}^3$ is the position of the robot, $^W\mathbf{q}_{I} \in \mathbb{H}$ is the quaternion over $\mathbb{S}^3$  in Hamiltonian representation, $^W\mathbf{v}_{I} \in \mathbb{R}^3$  is the speed of the robot, $\mathbf{b}_a \in \mathbb{R}^3$  is the bias of the accelerometer, $\mathbf{b}_{g} \in \mathbb{R}^3$ is the bias of the gyroscope. The IMU measurements $\mathbf{\hat a}$ and $\boldsymbol{\hat \omega}$ are modeled as
\begin{equation}
	\begin{aligned}
		 & \hat{\mathbf{a}}_i=\left(\mathbf{a}_i-\mathbf{g}\right)+\mathbf{b}_i^{\boldsymbol{a}}+\mathbf{n}_i^{\boldsymbol{a}} \\
		 & \hat{\boldsymbol{\omega}}_i=\boldsymbol{\omega}_i+\mathbf{b}_i^{\boldsymbol{g}}+\mathbf{n}_i^{\boldsymbol{g}},
		\label{eq_2_imu_model}
	\end{aligned}
\end{equation}
with $i=1,...,M$ index marks for $M$ measurements in $t_b^k$ and $t_e^k$ time intervals. The raw sensor measurements $\mathbf{a}_i$ and $\boldsymbol{\omega_i}$ contain bias $\mathbf{b}_i$ and white noise $\mathbf{n}_i$, and $\mathbf{g}$ is the rotated gravity vector. In this work, we address the following problem: between each frame received by LiDAR, given the cumulative point cloud $\mathcal{P}_k$ from LiDAR and the measurements $\mathbf{a}_i$ and $\boldsymbol{\omega}_i$ sampled by the IMU, estimate the robot's state $\mathbf{X}_b^k$ and $\mathbf{X}_e^k$ and the map $\mathcal{M}_k$. After optimization, we use the optimized pose to update the bias $\mathbf{b}_a$ and $\mathbf{b}_{g}$.

\subsection{Continuous-time based point-to-plane ICP}
When the IMU timestamp or the extrinsic parameter is not accurate, then using IMU is not up to the task of deskewing. So IMU is only used to estimate an initial pose. Based on this, we utilize the LO modality for deskewing.
Inspired by \cite{dellenbach2022ct}, for each frame of the point cloud $\mathcal{P}_k$ , we first extract the keypoints indexed by $\mathbf{I}_n$ indexed keypoints $\left\{\mathbf{p}_i\in\mathcal{P}_k|i\in\mathbf{I}_n\right\}$ , and then register the point cloud to the map. The map is composed of a point cloud registered to the world coordinate system $\mathcal{M}_n=\left\{\mathbf{p}_i^W\right\}$, which is stored in a voxel map consisting of multiple resolutions. The details of its construction are described in \ref{chap_7_adaptive_multiresolution_map}. We then optimize the first and last moments of this frame $t_{b}^i$, $t_{e}^i$ and the two states of this frame $\mathbf{X}_b^i$, $\mathbf{X}_e^i$ and transform the current frame point cloud $\mathcal{P}_k$ with the optimized pose to the world coordinate system.

These optimized poses $\mathbf{X}=\left\{\mathbf{X}_b^i,\mathbf{X}_e^i\right\}$ are solved by solving the following problem
\begin{equation}
	\underset{\mathbf{X}}{\arg \min } \{{r}_{\mathrm{icp}}(\mathbf{X})+\beta \mathbf{r}_c(\mathbf{X})\},
	\label{eq_3_ctmodel}
\end{equation}
where $\beta$ is the elastic weight. Position consistency constraints, orientation consistency constraints, and velocity consistency constraints are
\begin{equation}
	\mathbf{r}_c(\mathbf{\mathbf{X}})=\left[\begin{array}{c}
			\mathbf{r}_c^{\mathbf{q}}                  \\
			\mathbf{r}_c^{\mathbf{p}}                  \\
			\mathbf{r}_c^{\mathbf{v}}                  \\
			% \mathbf{r}_c^{\mathbf{b}_{\boldsymbol{g}}} \\
			% \mathbf{r}_c^{\mathbf{b}_{\boldsymbol{a}}}
		\end{array}\right]=\left[\begin{array}{c}
			2\left[\mathbf{q}_{e_i}^{W-1} \otimes \mathbf{q}_{b_{i+1}}^W\right]_{x y z} \\

			\mathbf{p}_{b_{i+1}}^W-\mathbf{p}_{e_i}^W                                   \\
			\mathbf{v}_{b_{i+1}}^W-\mathbf{v}_{e_i}^W                                   \\
			% \mathbf{b}_{\boldsymbol{g}_{b_{i+1}}}-\mathbf{b}_{\boldsymbol{g}_{e_i}}     \\
			% \mathbf{b}_{\mathbf{a}_{b_{i+1}}}-\mathbf{b}_{\mathbf{a}_{e_i}}             \\
		\end{array}\right],
	\label{eq_5_constrains}
\end{equation}
where $r_{icp}$ is the frame-to-map continuous time ICP
\begin{equation}
	r_{\mathrm{icp}}(\mathbf{\mathbf{X}})=\frac{1}{\left|\mathbf{I}^n\right|}\ \sum_{i\ \in \mathbf{I}^n}\ \rho\left(r_i^2(\mathbf{X})\right),
	\label{eq_4_residual}
\end{equation}
where the point cloud residual constraint ${\ r}_i$ is
\begin{equation}
	\begin{aligned}
		r_i(\mathbf{X})  & =\left(\mathbf{p}_i^W(\mathbf{X})-\mathbf{q}_i^W\right) \cdot \mathbf{n}_i                       \\
		\mathbf{p}_i^W   & =\mathbf{q}_{s_{i}}^W \mathbf{^Ip}_i+\mathbf{t}_{s_{i}}^W                                        \\
		\mathbf{q}_{s_i} & =\operatorname{slerp}\left(\mathbf{q}_{\mathbf{b}_i}, \mathbf{q}_{\mathbf{e}_i}, \alpha_i\right) \\
		\mathbf{p}_{s_i} & =\left(1-\alpha_i\right) \mathbf{p}_{\mathbf{b}_i}+\alpha_i \mathbf{p}_{\mathbf{e}_i}.
		\label{eq_6_constrains}
	\end{aligned}
\end{equation}

\section{System overview}\label{chap_4_system_overview}
\begin{figure}
	\centering
	\includegraphics[width=\columnwidth]{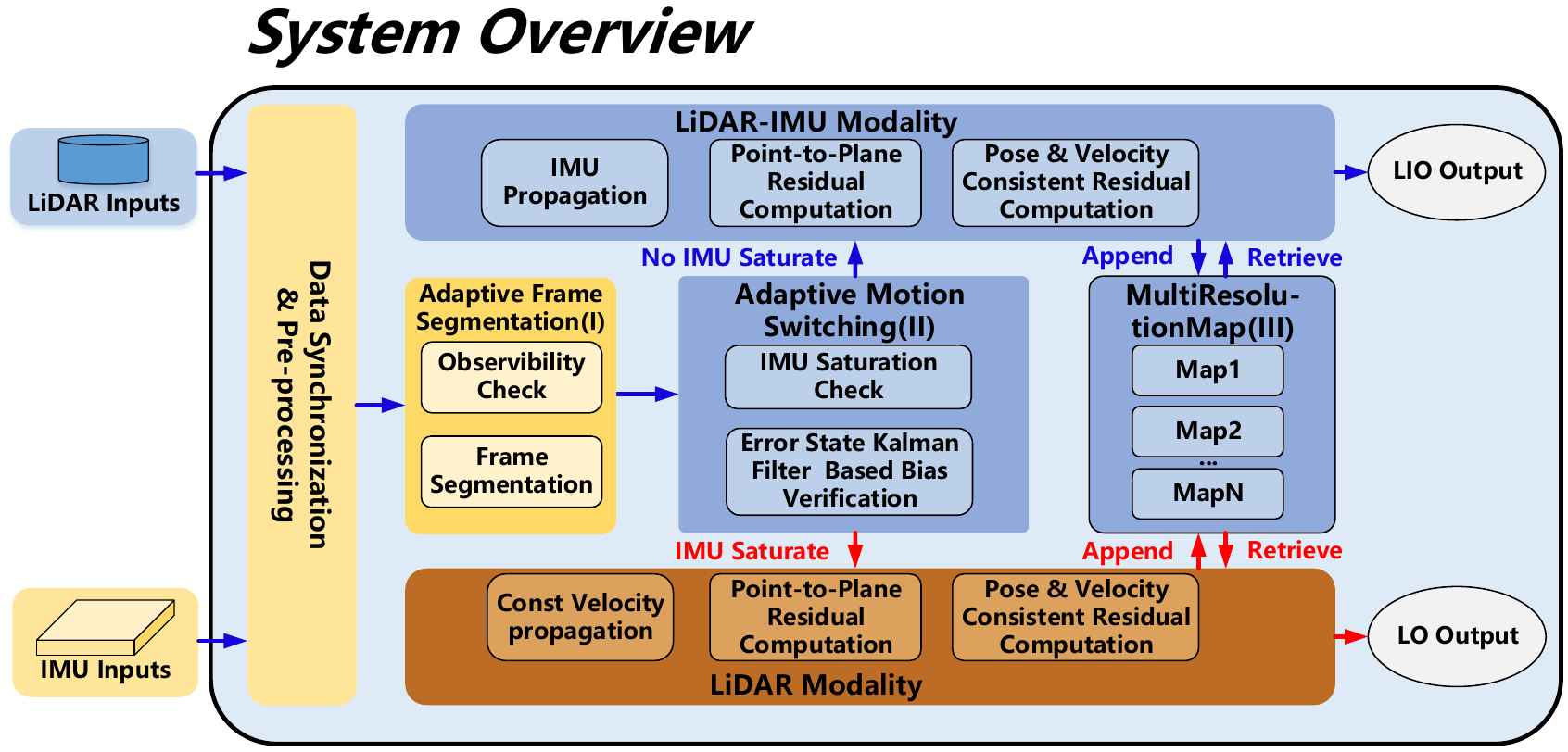}
	\caption{System pipeline.}
	\label{fig_1_system_overview2}
\end{figure}

Our goal is to estimate the LiDAR's 6-degree-of-freedom pose and simultaneously build a global map, and the system framework is outlined in Fig. \ref{fig_1_system_overview2}.
The whole system consists of three main modules:
\begin{itemize} % 使用 enumitem 包的 label 选项来改变标签样式
	\item \textbf{Adaptive Frame Segmentation Module:} Adopt adaptive frame segmentation method, combined with observability analysis, to determine whether to segment frame and how many frames should be segmented to improve the accuracy of the point cloud.
	\item \textbf{Adaptive Motion Modality Switching Module:} This module consists of an IMU saturation and fault detection and LIO switching module that performs back-end optimization and outputs initial odometry and undistorted features.
	\item \textbf{Adaptive Multi-Resolution Map Module:} This module constructs a multi-resolution point cloud map by updating the feature points, and the points in each frame are selected from the multi-resolution map set to construct the point-to-plane residual constraints, pose consistency constraints, and velocity consistency constraints to solve the optimal pose. Finally,  the points in the new frames are registered under the global coordinate system by the optimal pose.

\end{itemize}
% \begin{figure*}
% 	\centering
% 	\includegraphics[width=0.7\textwidth]{figs/fig_1_system_overview.pdf}
% 	\caption{System pipeline.}
% 	\label{fig_1_system_overview2}
% \end{figure*}

\section{Adaptive frame segmentation}\label{chap_5_adaptive_frame_segmentation}
The inputs to Adaptive-LIO are dense point cloud collected by LiDAR, along with linear acceleration and angular velocity measurements from a 6-axis IMU with synchronized timestamps. For devices with multi-sensor fusion for localization, achieving extrinsic calibration between sensors is often challenging, and implementing hardware time synchronization among various sensors can be costly. Therefore, we consider a loosely coupled model. However, using IMU data to undistort LiDAR points accurately under a loosely coupled model is challenging. Thus, we adopt a constant velocity motion model for undistortion in the loosely coupled scenario. Typically, for the constant velocity motion model, we assume a constant velocity period of 0.1s, which may differ from actual scenarios. The more frames we segment, the shorter the assumed constant velocity time, which better aligns with reality, leading to higher localization accuracy. However, frequency of frame segmentation must also consider the risk of degeneration. Therefore, we combine observability analysis to determine whether segmentation is necessary and how many frames should be segmented, ensuring both robustness and accuracy of the system. We proceed to elaborate on our adaptive frame segmentation strategy through two steps.

\textbf{Step 1}: We employ the observability analysis to determine whether to segment the frames and how many frames to segment.

For a point set $P=\{^W\mathbf{p}_1,..., ^W\mathbf{p}_i,..., ^W\mathbf{p}_m\}$, we calculate the point-to-plane residual $r_i$  for each point $^W\mathbf{p}_i$ using \ref{eq_6_constrains}, and then compute the Jacobian matrix $\mathbf{J}_i$ of $r_i$ with respect to the pose $\mathbf{T}=\left\{\mathbf{T}_b,\mathbf{T}_e\right\}$, where $\mathbf{T}_b=\left\{\mathbf{R}_b,\mathbf{p}_b\right\}$ and $\mathbf{T}_e=\left\{\mathbf{R}_e,\mathbf{p}_e\right\}$.
\begin{equation}
	\begin{split}
		\mathbf{J}_i &= \frac{\partial r_i}{\partial \mathbf{T}} = \frac{\partial r_i}{\partial\left[\begin{array}{llll}
					\mathbf{R}_b & \mathbf{p}_b & \mathbf{R}_e & \mathbf{p}_e
				\end{array}\right]} \\
		&= \left[\begin{array}{cccc}
				\frac{\partial r_i}{\partial \mathbf{R}_b} & \frac{\partial r_i}{\partial \mathbf{p}_b} & \frac{\partial r_i}{\partial \mathbf{R}_e} & \frac{\partial r_i}{\partial \mathbf{p}_e}
			\end{array}\right]_{1 \times 12}.
	\end{split}
	\label{eq_7_Ji}
\end{equation}

We concatenate the Jacobian matrices $\mathbf{J}_i$ of each point to obtain the total Jacobian matrix $\mathbf{J}$
\begin{equation}
	\mathbf{J}=\left[\begin{array}{lllll}
			\mathbf{J}_1^T & \cdots & \mathbf{J}_i^T & \cdots & \mathbf{J}_m^T
		\end{array}\right]_{m \times 12}^T,
	\label{eq_8_J}
\end{equation}
let $\mathbf{A}=\mathbf{J}^T \mathbf{J}$, which represents the information matrix of rotations and translations
\begin{equation}
	\mathbf{A}=\mathbf{J}^T\mathbf{J}
	= \left[\begin{array}{cccc}

			\mathbf{A}_{R_b} & \cdots           & \cdots           & \cdots           \\

			\vdots           & \mathbf{A}_{p_b} & \cdots           & \cdots           \\

			\vdots           & \vdots           & \mathbf{A}_{R_e} & \cdots           \\

			\vdots           & \vdots           & \cdots           & \mathbf{A}_{p_e} \\
		\end{array}\right]_{12 \times 12}, \\
	\label{eq_9_A}
\end{equation}
where $\mathbf{A}_{R_b},\mathbf{A}_{p_b},\mathbf{A}_{R_e},\mathbf{A}_{p_e}$ represent the information sub-matrices for the rotation and translation of the initial and final frames, respectively. We perform Singular Value Decomposition (SVD) on them to detect degradation. The degeneration detection steps are as follows:

(1) Firstly, we perform SVD on the $\mathbf{A}_{R_b},\mathbf{A}_{p_b},\mathbf{A}_{R_e},\mathbf{A}_{p_e}$ to obtain the eigenvalues respectively
$\boldsymbol{\lambda}_{\mathbf{R}_b}=\{\lambda_{\mathbf{R}_b}^1,\lambda_{\mathbf{R}_b}^2,\lambda_{\mathbf{R}_b}^3 \}$, $\boldsymbol{\lambda}_{\mathbf{p}_b}=\{\lambda_{\mathbf{p}_b}^1,\lambda_{\mathbf{p}_b}^2,\lambda_{\mathbf{p}_b}^3 \}$, $\boldsymbol{\lambda}_{\mathbf{R}_e}=\{\lambda_{\mathbf{R}_e}^1,\lambda_{\mathbf{R}_e}^2,\lambda_{\mathbf{R}_e}^3 \}$, $\boldsymbol{\lambda}_{\mathbf{p}_e}=\{\lambda_{\mathbf{p}_e}^1,\lambda_{\mathbf{p}_e}^2,\lambda_{\mathbf{p}_e}^3 \}$, where $\lambda_{(.)}^1 \geq \lambda_{(.)}^2 \geq \lambda_{(.)}^3$.

(2) Second, we set 2 constants $\epsilon_R$,$\epsilon_p$ to be the degeneration judgment thresholds for the rotational and translational components respectively, we define $\alpha_R$, $\alpha_p$ to represent the degeneration factors for rotation and translation, respectively, as follows:
\begin{equation} \alpha_{\mathbf{R}_b}=\frac{\lambda_{\mathbf{R}_b}^3}{\lambda_\mathbf{{R}_b}^2},\alpha_{\mathbf{p}_b}=\frac{\lambda_{\mathbf{p}_b}^3}{\lambda_{\mathbf{p}_b}^2},\alpha_{\mathbf{R}_e}=\frac{\lambda_{\mathbf{R}_e}^3}{\lambda_{\mathbf{R}_e}^2},\alpha_{\mathbf{p}_e}=\frac{\lambda_{\mathbf{p}_e}^3}{\lambda_{\mathbf{p}_e}^2}.
	\label{eq_10_factor}
\end{equation}

A larger $\alpha$ indicates weaker observability, and thus fewer frames should be segmented, and vice versa. We determine whether a frame has degenerate by comparing the sizes of $\alpha$ and $\epsilon$. If $\alpha$ satisfies any of the following inequalities
\begin{equation}
	\begin{cases}
		\alpha_{\mathbf{R}_b},\alpha_{\mathbf{R}_e} > \epsilon_\mathbf{R} \\
		\alpha_{\mathbf{p}_b},\alpha_{\mathbf{p}_e} > \epsilon_\mathbf{p}
	\end{cases},
	\label{eq_11_observability_alpha}
\end{equation}
then the observability of that frame is considered weak, and no frame segmentation is performed; otherwise, frame segmentation is conducted.

(3) Finally, we define a mapping function $d$ to determine the number of segmented frames. Under the condition of frame segmentation, we first define a total degeneration factor $\alpha=\frac{\alpha_{\mathbf{R}_b}+\alpha_{\mathbf{p}_b}+\alpha_{\mathbf{R}_e}+\alpha_{\mathbf{p}_e}}{4}$,  and then map $\alpha$ to $d$ as follows
\begin{equation}
	d = d_{min}(1-\alpha)+d_{max} \alpha,
	\label{eq_12_d_map}
\end{equation}
where  $d_{min}=1$, $d_{max}=4$, representing the minimum and maximum number of frames that can be segmented from the original frame, respectively. Through observability analysis, we achieve adaptive frame segmentation.

\textbf{Step 2:} Based on the $d$, we will elaborate on how to segment a frame of LiDAR point cloud $d$ into frames in a specific process.

\begin{figure}
	\centering
	\includegraphics[width=1\columnwidth]{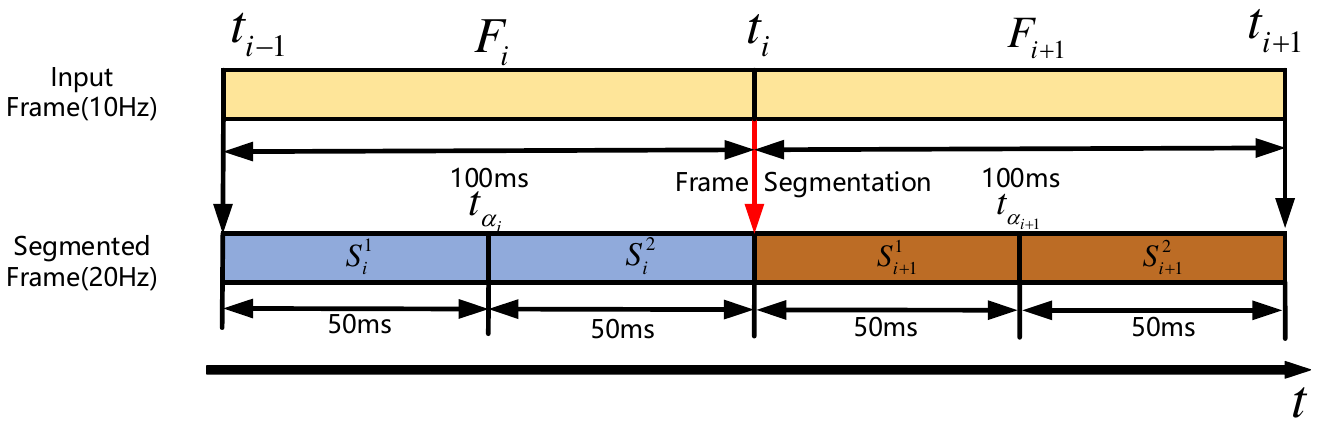}
	\caption{Frame segmentation strategy.}
	\label{fig_2_frame_segmentation}
\end{figure}

Frame segmentation aims to divide a longer LiDAR intervals from a particular frame into multiple segments of shorter time intervals. As an example, we describe how we output a 20 Hz point cloud $\mathcal{P}$ from a 10 Hz raw input point cloud $F$, we present our strategy of frame segmentation as shown in Fig. \ref{fig_2_frame_segmentation}.

Assuming that the last frame $F_i$ starts at $t_{i-1}$ and ends at $t_i$ and the current frame $F_{i+1}$ begins at $t_i$ and ends at $t_{i+1}$. We assume that the time intervals $[t_{i-1}, t_i]$ and $[t_i, t_{i+1}]$ are both 100ms in length. We first sort the acquired point clouds in chronological order from smallest to largest, based on the continuous acquisition characteristics of LiDAR over a period of time, we can segment the original frames into consecutive point cloud data, and then re-package the point cloud data to obtain frames of higher frequency. Specifically, we first calculate the midpoints of the time intervals $[t_{i-1}, t_i]$ and $[t_i, t_{i+1}]$ (i.e. $t_{\alpha_i}$ and $t_{\alpha_{i+1}}$), and put all the timestamps into a set
\begin{equation}
	N= \{t_{{i-1}}, t_{\alpha_i}, t_i,t_{\alpha_{i+1}},t_{i+1}\}.
	\label{eq_13_N}
\end{equation}

We take each element of N (i.e.$N\left[l\right]$ ) as the starting timestamp and $N\left[l+2\right]$  as the ending timestamp. Ultimately, we reduce the time interval between the two frames (i.e.$S_i^1$ and $S_i^2$ )  from 100 milliseconds to 50 milliseconds. Therefore, our proposed frame segmentation method can decrease the time interval and improve the accuracy of the point cloud.

\textbf{Remark:} Our frame segmentation strategy is designed for non-repetitive solid-state scanning LiDAR. It may not be effective for conventional rotating LiDARs, so generally, frame segmentation is not recommended. The \textbf{Adaptive Motion Modality Switching Module} and the \textbf{Adaptive Multi-Resolution Map Module} are equally applicable to both mechanical LiDAR and non-repetitive solid-state scanning LiDAR.

\section{Adaptive motion modality switching module}\label{chap_6_adaptive_motion_switching}
By conducting saturation and fault detection on the IMU state, we can determine whether to utilize LO modality or LIO modality for the subsequent step. LO modality employs a constant velocity model to compute the state of the next frame, whereas LIO modality utilizes IMU data to compute the state of the next frame. We divide this process into two steps: the first step involves saturation and fault detection of the IMU, while the second step focuses on the inter-switching of LO and LIO modality, which will describe in detail below.

\textbf{Step 1: IMU saturation and fault detection}

IMUs play an important role in LIO systems, but in practice, IMUs can have many problems, such as IMU timestamp error, IMU over-range, and so on. These problems affect the robustness of the SLAM system, for this reason we need to perform saturation and fault detection on the IMU to facilitate adaptive switching between LO and LIO modality. Our adaptive switching method is to use LO modality if the IMU exceeds the range, and use LIO modality if the IMU does not exceed the range.

\textbf{Step 2: Mutual switching of LO and LIO}

\begin{figure}
	\centering
	\includegraphics[width=1.05\columnwidth]{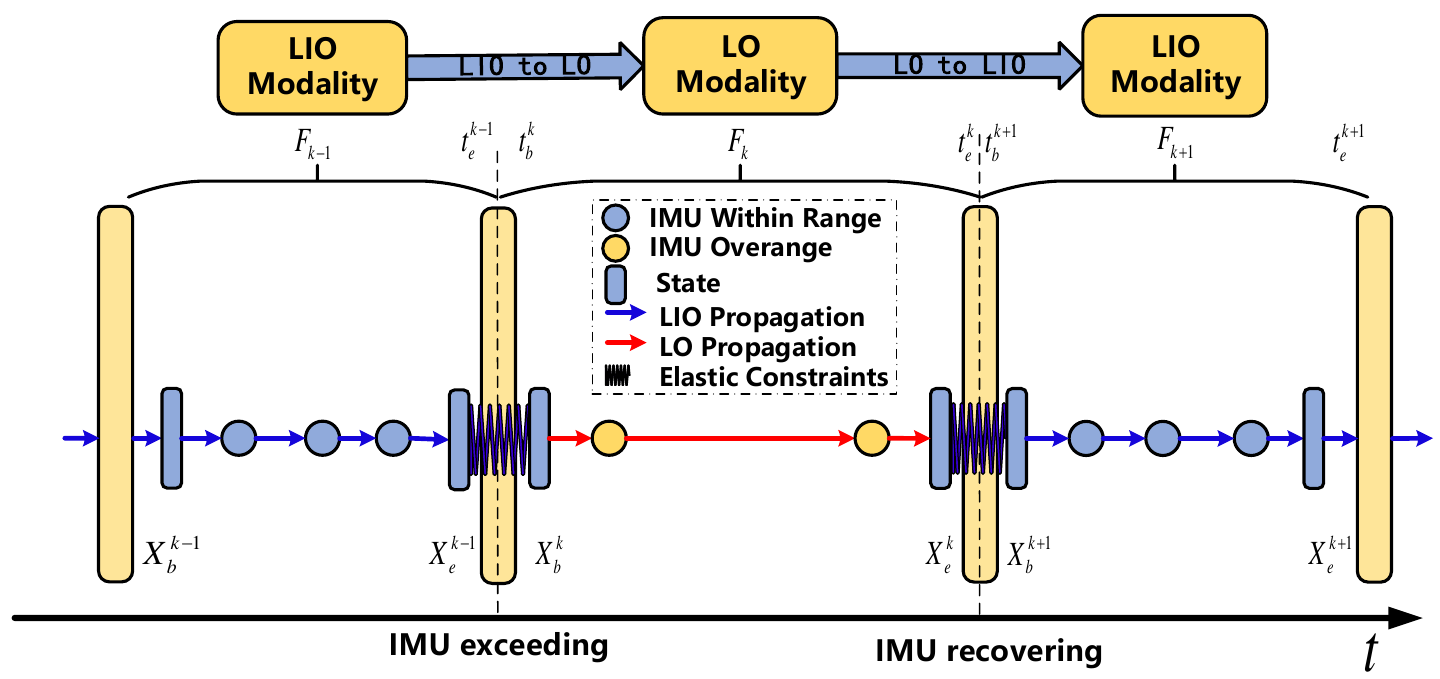}
	\caption{LO and LIO switching.}
	\label{fig_3_liolo_switching}
\end{figure}

We define a sequence of consecutive LiDAR frames as $\mathbf{F}=\{...,F_{k-1},F_k,F_{k+1},...\}$, as illustrated in the Fig. \ref{fig_3_liolo_switching}. For each frame $F_i$, its begin and end states are denoted as $\mathbf{X}_{b}^i,\mathbf{X}_{e}^i$, respectively, with corresponding timestamps $t_b^i,t_e^i$ ($i=\{...,k-1,k,k+1,...\})$. Additionally, the timestamp of the end of the previous frame is equal to the timestamp of the start of the next frame, i.e., $t_b^{i-1}=t_e^{i}$.
We determine whether to switch to LO modality or LIO modality based on whether the IMU has exceeded its range. At time $t=t_b^{k-1}$, the IMU is within range, and the system propagates the state in LIO modality within the interval $t\in(t_b^{k-1},t_b^k)$(refer to \ref{chap_61_lio2lo}); at time $t=t_b^k$, the IMU exceeds its range, prompting the system to switch from LIO modality to LO modality, and the state is propagated in const velocity within the interval $t\in(t_b^{k},t_b^{k+1})$; at time $t=t_b^{k+1}$, the IMU returns to the normal state, and the system evaluates the convergence of $b_a$,$b_g$ to determine whether to switch back to LIO modality(refer to \ref{chap_62_lo2lio}); in the $t=t_b^{k+1}$, the system continuously propagates the state in LIO modality.

\subsection{LIO mode switched to LO mode}\label{chap_61_lio2lo}
When each downsampling frame $F_{k-1}$ is completed, we use IMU to predict the prior values of the state at the beginning of $F_{k-1}$, denoted as  ${\hat{\mathbf{X}}}_b^{k-1}$, and at the end of $F_{k-1}$ , denoted as ${\hat{\mathbf{X}}}_e^{k-1}$, for LIO optimization. Specifically, the predicted states $\hat{\mathbf{X}}_e^{k-1}$ are:
\begin{equation}
	\mathbf{\hat {X}}_b^{k-1}=\mathbf{\overline{X}}_e^{k-2},
	\label{eq_14_predicted_state}
\end{equation}
specifically, the predicted states ${\hat{\mathbf{X}}}_e^{k-1}$ are
\begin{equation}
	\begin{aligned}
		\boldsymbol{\hat{R}}_{e}^{k-1} & = \boldsymbol{\hat{R}}_i^{k-1} \prod_{i=1}^{n} \operatorname{Exp}\left(\left(\boldsymbol{\omega}_i - \boldsymbol{\hat{b}}_{g_i}^{k-1} - \boldsymbol{n}_{g_i}\right) \delta t\right),                                      \\
		\boldsymbol{\hat{v}}_e^{k-1}   & = \boldsymbol{\hat{v}}_i^{k-1} + \boldsymbol{g}^W \sum_{i=1}^n \delta t + \sum_{i=1}^{n} \boldsymbol{\hat{R}}_{i}^{k-1} \left(\boldsymbol{a}_i - \boldsymbol{\hat{b}}_{a_i}^{k-1} - \boldsymbol{n}_{a_i}\right) \delta t, \\
		\boldsymbol{\hat{p}}_e^{k-1}   & = \boldsymbol{\hat{p}}_i^{k-1} + \sum_{i=1}^{n} \boldsymbol{\hat{v}}_i^{k-1} \delta t + \frac{1}{2} \sum_{i=1}^{n} \boldsymbol{g}^W \delta t^2                                                                            \\
		                               & + \frac{1}{2} \sum_{i=1}^{n} \boldsymbol{\hat{R}}_i^{k-1} \left(\boldsymbol{a}_i - \boldsymbol{\hat{b}}_{a_i}^{k-1} - \boldsymbol{n}_{a_i}\right) \delta t^2,
	\end{aligned}
	\label{eq_15_imu_state}
\end{equation}
where $\mathbf{g}^W$ is the gravity in the world coordinate system, $i$ and $i+1$ are the two moments when IMU measurements are obtained during $[t_{e}^{k-2},t_{e}^{k-1}]$, $\delta t$ is the time interval between $i$ and $i+1$, $n = \left(t_{e}^{k-1} - t_{e}^{k-2} \right) / \delta t$, and when $n = 1$, $\mathbf{\hat X}_n^{k-1}=\mathbf{\overline X}_{b}^{k-2}$.
For $\mathbf{b}_{a_{e}}^{k-1}$ and $\mathbf{b}_{g_e}^{k-1}$, we set their predicted values as follows $\mathbf{b}_{a_{e}}^{k-1} = \mathbf{b}_{a_{e}}^{k-2}$ and $\mathbf{b}_{g_{e}}^{k-1} = \mathbf{b}_{g_{e}}^{k-2}$.

We obtain the optimized states of the head and tail of this frame $F_{k-1}$  through (\ref{eq_5_constrains}), denoted as $\mathbf{\check X}_{k-1}=(\mathbf{\check X}^{k-1}_b, \mathbf{\check X}^{k-1}_e)$, then we use the optimized tail state of this frame to update the head state of next frame, i.e., $\mathbf{X}_b^k=\mathbf{\check X}_{e}^{k-1}$.

Since at the $t_b^k$, after IMU saturation detection, the IMU is out of range, and we use the LO modality as the propagation state for the next frame of the LO modality $\mathbf{\hat X}_b^k$
\begin{equation}
	\begin{aligned}
		\mathbf{\hat X}_b^k=[\mathbf{ R}_b^k,\mathbf{p}_b^k,\mathbf{v}_{const}^k,\mathbf{\hat b}_g^k, \mathbf{\hat b}_a^k]
		%\mathbf{\hat b}_a^k, \mathbf{\hat g}^k],
	\end{aligned}
	\label{eq_16_nominal_state}
\end{equation}
among them
\begin{equation}
	\begin{aligned}
		\mathbf{v}_{const}^k & = \frac{\mathbf{p}_{e}^{k-1}-\mathbf{p}_{b}^{k-1}}{\mathbf{t}_{e}^{k-1}-\mathbf{t}_{b}^{k-1}}.
	\end{aligned}
	\label{eq_17_velocity}
\end{equation}

Eventually we completed the switch from LIO to LO modality.

\subsection{LO mode switching to LIO mode}\label{chap_62_lo2lio}

After switching back to LO mode, we obtain the state $\mathbf{\hat X}_{b}^k$ at the beginning of the frame $F_k$. However, since the IMU is saturated at this time, we get the state at the end of the frame by using the LO modality
\begin{equation}
	\begin{aligned}
		\mathbf{\hat R}_{e}^k & =\mathbf{\check R}_{e}^{k-1} \mathbf{\check R}_{b}^{^{-1}k-1} \mathbf{\hat R}_{b}^k                                                      \\
		\mathbf{\hat p}_{e}^k & = \mathbf{\hat p}_{b}^k+\mathbf{\check R}_{e}^{k-1} \mathbf{\check R}_{b}^{^{-1}k-1}(\mathbf{\hat p}_{b}^k-\mathbf{\check p}_{b}^{k-1}). \\
	\end{aligned}
	\label{eq_18_lo2lio}
\end{equation}

By repeatedly applying (\ref{eq_5_constrains}), we optimize the pose of the frame to obtain $\mathbf{\check X}_{k}=(\mathbf{\check X}_b^{k},\mathbf{\check X}_e^{k})$, and then we use the optimized end-of-frame state to update the true state of the begin of the next frame, i.e.$\mathbf{X}_b^{k+1}=\mathbf{\check X}_{e}^{k}$.

At the moment $t_e^k (t_b^{k+1})$, we perform saturation detection on the IMU, and find that the IMU is not out of range. We then determine whether to switch to LIO mode or maintain LO mode unchanged by assessing the convergence of the estimated bias $\mathbf{\hat b}_{a_b}^{k+1},\mathbf{\hat b}_{g_b}^{k+1}$ of initial IMU states for the next frame $F_{k+1}$. The following are our steps:

\textbf{Step 1:} Define the Propagation state $\mathbf{\hat X}_e^{k}$ at the end of the frame $F_k$ to be the prediction state $h(\mathbf{\hat X}_e^{k})=(\mathbf{\hat R}_e^k,\mathbf{\hat p}_e^k)$. The state $\mathbf{\hat X}_e^{k}$ is iteratively obtained by (\ref{eq_15_imu_state}) ($n=\left(t_{e}^{k}-t_{b}^{k}\right) / \delta t$), and the optimized state $\mathbf{z}_e^k=(\mathbf{\check R}_e^{k},\mathbf{\check p}_{e}^{k})$) at the end of this frame is considered as the observation. We update it using the ESKF as follows
\begin{equation}
	\begin{aligned}
		\delta \mathbf{\hat X}_e^k & = K (\mathbf{z}_e^k \boxminus h(\mathbf{\hat X}_e^k))       \\
		\mathbf{\overline X}_e^{k} & = \mathbf{\hat X}_e^{k}\boxplus \delta \mathbf{\hat X}_e^k,
	\end{aligned}
	\label{eq_19_eskf_update}
\end{equation}
where $K$ is the Kalman gain and $\boxplus,\boxminus$ are the generalized addition and generalized subtraction, whose algorithms are referred to \cite{xu2022fast}. We obtain the updated accelerometer and gyroscope biases $\mathbf{\overline b}_{a_e}^{k},\mathbf{\overline b}_{g_e}^{k} \in \mathbf{\overline X}_e^{k}$, which are then used to switch the motion modality in the next step. \textbf{Note:} IMU data only provides an initial value for state propagation in LIO modality and does not directly update the system state variables.

\textbf{Step 2:} We define a threshold $\epsilon$ ,if $(\mathbf{\overline b}_{a_e}^{k},\mathbf{\overline b}_{g_e}^{k}) \leq \epsilon$, then it switches to LIO modality and we update the propagated state of the next frame $F_{k+1}$  with $\mathbf{\overline X}_e^{k}$ as follows
\begin{equation}
	\begin{aligned}
		\mathbf{\hat X}_b^{k+1}=\mathbf{\overline X}_e^{k},
	\end{aligned}
	\label{eq_20_update}
\end{equation}
among them
\begin{equation}
	\begin{aligned}
		\mathbf{\hat X}_b^{k+1}=[\mathbf{R}_b^{k+1},\mathbf{p}_b^{k+1},\mathbf{v}_{const}^{k+1}, \mathbf{\hat b}_g^{k+1}, \mathbf{\hat b}_a^{k+1}].
		%		\mathbf{\hat g}^{k+1}
	\end{aligned}
	\label{eq_21_update_bias}
\end{equation}

We finalized the switch from LO to LIO.

\section{Adaptive multi-resolution map module}\label{chap_7_adaptive_multiresolution_map}
Regardless of changes in the environment, fixing the size voxels may result in maps with inconsistent density. Voxelization represents some points in the same voxel as individual points, so that there are fewer points in open areas than in narrow areas. Specifically, empty areas are larger and have sparser point clouds, while in narrow areas they are smaller and have denser point clouds. If multi-resolution voxel maps are used, i.e., high-resolution voxel maps in open areas and low-resolution voxel maps in narrow areas, the map accuracy can be improved and more constraints can be constructed in open areas, which ultimately can weaken the effects of degeneration to a certain extent.

In this section, we describe the construction of multi-resolution maps and the selection of adaptive multi-resolution maps.

\subsection{Map structure}
\begin{figure}
	\centering
	\includegraphics[width=.9\columnwidth]{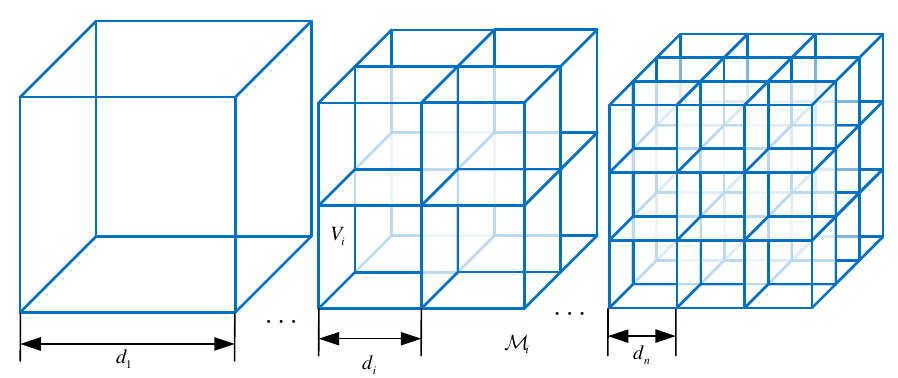}
	\caption{Multi-resolution voxel map. voxel $i$ contains a set of points $\mathbf{V}_i$ , and each point in $\mathbf{V}_i$ defines a nearest neighbor, where the relationship between points, voxels and maps is denoted as $\mathbf{p}_i \in \mathbf{V}_i\subset \mathbf{\mathcal{M}}_i, \mathbf{\mathcal{M}}_i \subset \mathcal{M}$.}
	\label{fig_4_multi_resolution_map}
\end{figure}

We construct a multi-resolution voxel map organized by hash tables $\mathcal{M}=\{\mathcal{M}_1,...,\mathcal{M}_n\}$. As shown in Fig. \ref{fig_4_multi_resolution_map}, each map $\mathcal{M}_i$ has four features, voxel $V$ , resolution $d$, maximum number of points $n$ and maximum neighborhood radius $R$ ,i.e.,$\mathcal{M}_i=(V_i,d_i,n_i,R_i)$. Define a set of resolution sets $F = \{d_1,...,d_i,...,d_n\}$, where $d_1\leq ...\leq d_i \leq,...,\leq d_n$, and $i$ is the map id and the resolution $d_i$ corresponds to the map $\mathcal{M}_i$. We establish a mapping function to insert point $\mathbf{p}_i$ into voxel maps of different resolutions and find the nearest neighbors. Below, we will describe our strategy.

\subsection{Adaptive resolution map selection}
\begin{figure}
	\centering
	\includegraphics[width=.9\columnwidth]{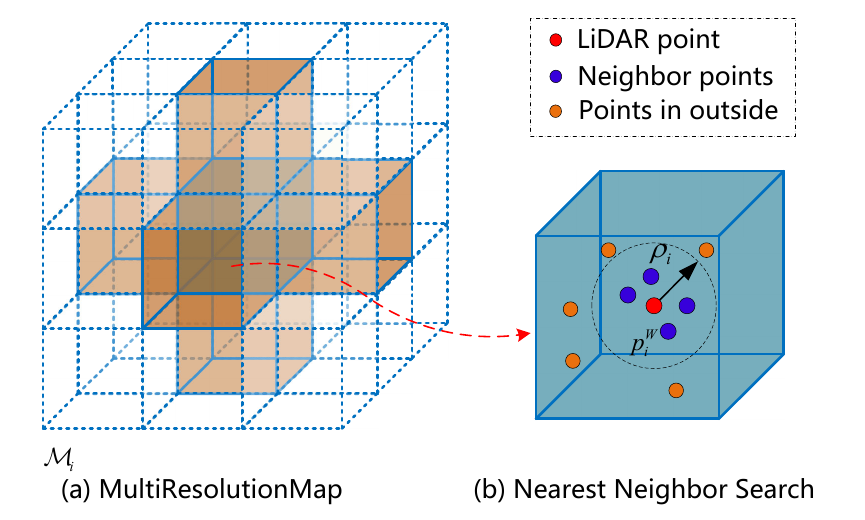}
	\caption{Nearest neighbor search. Finding a given point $\mathbf{p}_i$ of the neighboring voxels, and then search through the radius to get the nearest neighbors $\left\{\mathbf{q}_i^W\right\}$ of a given point $\mathbf{p}_i$. Here we set the number of nearest neighbor voxels to 6.}
	\label{fig_5_nearest_search}
\end{figure}

As shown in Fig.\ref{fig_5_nearest_search}, Our approach is to adaptively set the size of the map resolution by checking whether the observed surroundings are narrow scenes. Given a point $\mathbf{p}_i$ , we will go through 4 steps on how to add points to a voxel map with different resolutions:

1) First, with $\|^L\mathbf{p}_i\|$ as the radius (i.e., the distance of $\mathbf{p}_i$ from the LiDAR system $L$ as the radius), interpolation is used to calculate the search radius $\rho_i$ as follows
\begin{equation}
	\rho_i = \beta r_{max}+(1-\beta)r_{min},
	\label{eq_22_search_radius}
\end{equation}
where $\beta$ represents the exponent determining the search radius based on the distance of the sensor, and the $r_{min}$ and $r_{max}$ represent the maximum and minimum radius of $\mathbf{p}_i$ from the center of the LiDAR coordinate system $L$. We use linear interpolation to calculate the radius between $r_{min}$ and $r_{max}$, allowing the neighborhood radius to dynamically adjust with distance changes.

2) We traverse through each resolution in $F$, finding the first resolution $d_i$  that is not less than $\rho_i$. Then, we index the corresponding voxel hash map $\mathcal{M}_i$ using the map ID $i$.

3) Calculate the current point $\mathbf{p}_i$ of the voxel corresponding to the voxel coordinates $\mathbf{v}_i=\frac{\mathbf{p}_i}{d_i}$ , we use $\mathbf{v}_i$ to index $\mathbf{p}_i$ corresponding voxel $V_i$ and then traverse $V_i$ over on 6 voxels centered at the top, bottom, front, back, left and right of the voxels $\{V_{i_1},...,V_{i_6}\}$, and find the nearest neighbors $\{\mathbf{q}_i^W\}$ among these voxels in these voxels.

4) We construct a neighboring plane $\boldsymbol{\pi}_i$  around $\mathbf{p}_i$ with the nearest neighbors $\{\mathbf{q}_i^W\}$ and calculate the point-plane residual $r_i$ of $\mathbf{p}_i$ from the neighboring plane. Readers can refer to (\ref{eq_4_residual}) for specific methods of constructing the point-plane residual.

\section{Experiment}\label{chap_8_experiment}

To demonstrate the outstanding performance of Adaptive-LIO, extensive experiments were conducted on public datasets and in real-world environments. The results show that Adaptive-LIO achieves superior accuracy and real-time performance compared to  other state-of-the-art LiDAR odometry algorithms.  We selected DLIO, LIO-SAM, Point-LIO, FAST-LIO2, and IG-LIO as comparison, all of which have demonstrated outstanding performance in the field of LiDAR odometry. All experiments were conducted in a Robot Operating System( ROS ) using an Intel i5-12700H CPU and 16GB of RAM, with all algorithms tested under the same parameters.

\begin{figure}[h] %h表示在该位置，htpn表示浮动在任何位置，不推荐
	\centering
	\includegraphics[width=0.7\columnwidth]{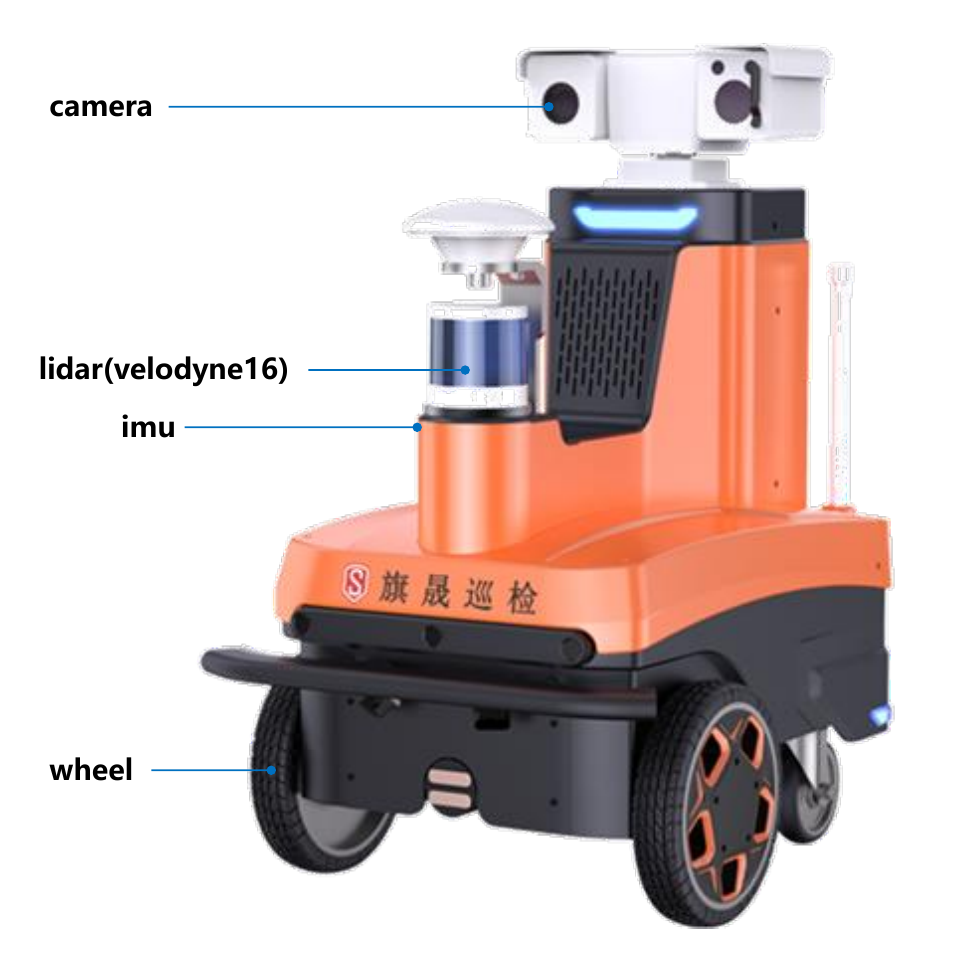}
	\caption{The experimental platform is the Qisheng L1 mobile chassis, which features dual-wheel differential steering. It is equipped with a Velodyne VLP-16 and an external IMU, using an Intel Core i5 as the computing platform. Please note that our IMU and LiDAR have not undergone hardware time synchronization, and the extrinsics between the LiDAR and IMU have not been strictly calibrated.}
	\label{fig_1_qishengrobot}
\end{figure}

\subsection{Extrinsic calibration and hardware time synchronization}\label{chap_81_extrinsic_calibration}
\begin{figure}
	\centering
	\includegraphics[width=0.5\textwidth]{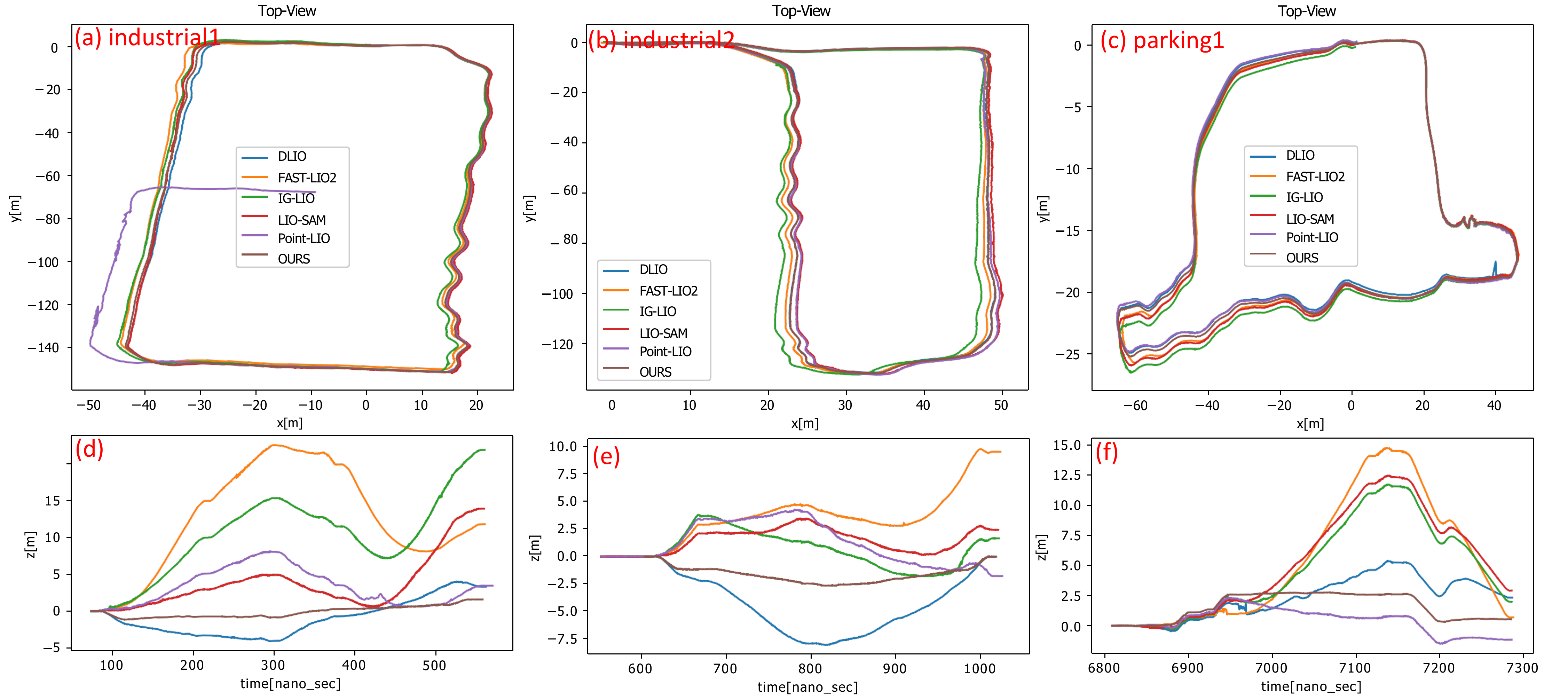}
	\caption{The comparison of Adaptive-LIO with other algorithms on datasets without hardware time synchronization or extrinsic calibration. Figures (a)-(c) and (d)-(f) show the 2D trajectory plots and the Z-axis trajectory over time for all algorithms on the datasets industrial1, industrial2, and parking1, respectively.}
	\label{fig_1_endtoend_traj}
\end{figure}

\begin{figure}
	\centering
	\includegraphics[width=0.5\textwidth]{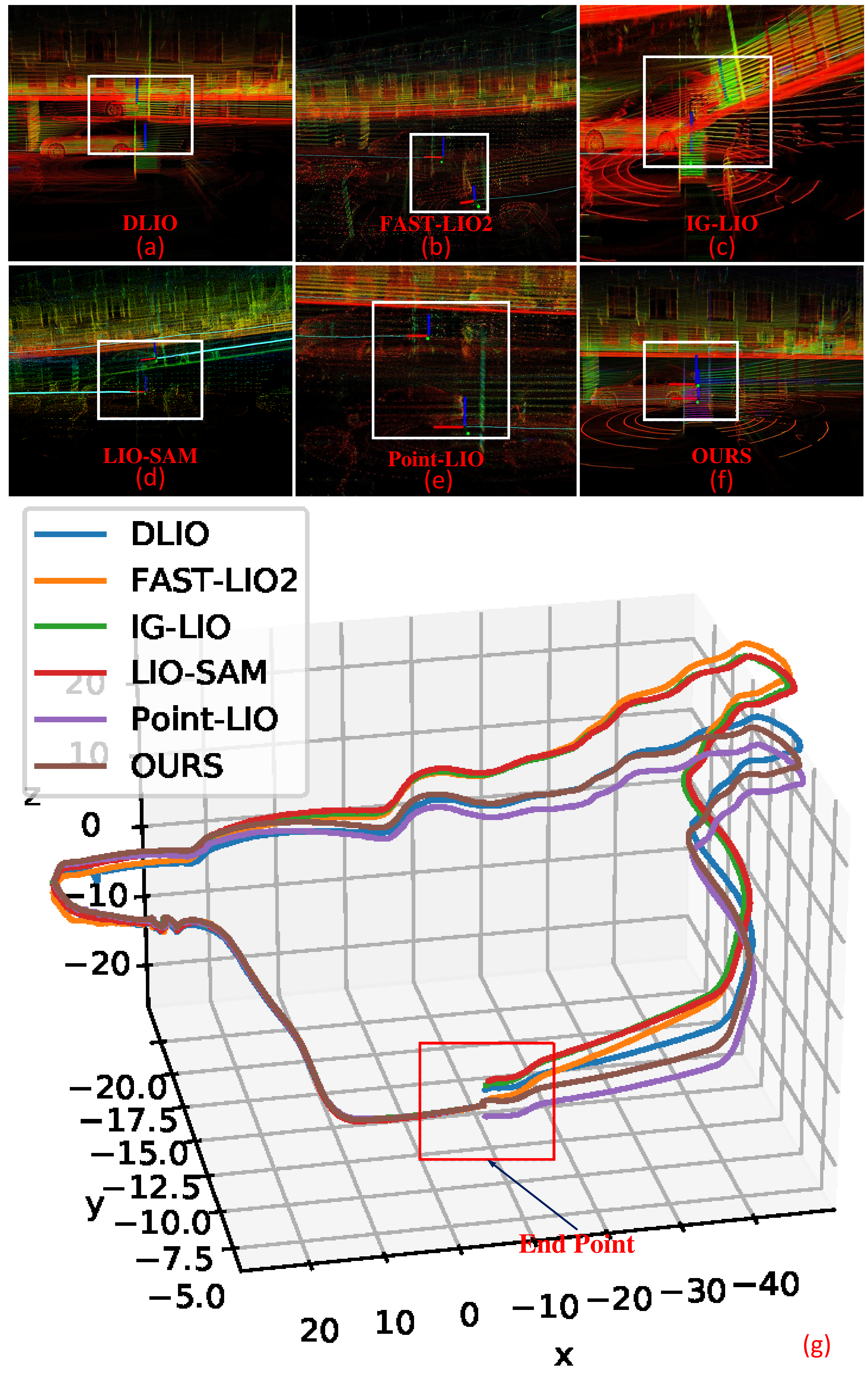}
	\caption{The performance of all algorithms on the parking1 dataset is illustrated in Figure (a)-(g). These figures show the mapping results of DLIO, LIO-SAM, Point-LIO, FAST-LIO2, IG-LIO, and our method (OURS) on the parking1 dataset. When the LiDAR returns near the original starting point, at the "End Point" location shown in Figure (g), significant Z-axis errors are observed in (a) DLIO, (b) FAST-LIO2, (c) IG-LIO, (d) LIO-SAM, and (e) Point-LIO. In contrast, (f) Adaptive-LIO demonstrates much smaller Z-axis errors.}
	\label{fig_1_endtoend_mapping}
\end{figure}
To evaluate the accuracy, we tested the performance of Adaptive-LIO on the Qisheng L1 mobile platform, as shown in Fig. \ref{fig_1_qishengrobot}, which is equipped with a VLP-16 LiDAR and an external IMU. It is important to note that the LiDAR and IMU were not hardware time-synchronized, and the extrinsic parameters were not strictly calibrated.

We recorded three datasets—named Qisheng dataset-industrial1, industrial2, and parking1—to test the performance of Adaptive-LIO. Fig. \ref{fig_1_endtoend_traj} shows the comparison results of our algorithm with other algorithms on datasets without hardware time synchronization and extrinsic calibration. It can be observed that our algorithm exhibits the least fluctuation in the Z-axis across all three datasets, while other algorithms show greater fluctuations in the Z-axis. This is because Adaptive-LIO is a loosely coupled algorithm, which better adapts to data without hardware time synchronization.

Fig. \ref{fig_1_endtoend_mapping}(a)-(f) illustrate the performance of each algorithm on the parking1 dataset, and Fig. \ref{fig_1_endtoend_mapping}(g) shows the 3D trajectories. At the endpoint, significant Z-axis errors were observed in DLIO, FAST-LIO2, IG-LIO, LIO-SAM and Point-LIO. In contrast, Adaptive-LIO exhibited much smaller Z-axis errors. We used end-to-end error as the evaluation metric, with results presented in Table \ref{tbl_chap_81_endtoend}. Adaptive-LIO performed the best across all three sequences, minimizing end-to-end error. In comparison, the tightly coupled framework FAST-LIO2 performed the worst on all three datasets, while Point-LIO failed on the industrial1 dataset, ultimately calculating a larger end-to-end error than Adaptive-LIO.

\begin{table}[htbp]
    \centering
    \caption{END TO END ERRORS (METERS) IN QISHENG DATASET.}\label{tbl_chap_81_endtoend}
    \begin{threeparttable}
        \begin{tabularx}{\linewidth}{@{}l *{6}{X}@{}}
            \toprule
            Dataset & DLIO & LIO-SAM & Point-LIO & FAST-LIO2 & IG-LIO & Ours \\
            \midrule
            industrial1 & 4.485 & 13.935 & ${}^{a}$ $\times$ & 11.778 & 21.815 & $^{b}$\textbf{2.4824} \\
            industrial2 & 0.185 & 2.467 & 1.778 & 9.547 & 1.737 & \textbf{0.107} \\
            parking1 & 1.81 & 2.27 & 3.164 & 5.53 & 1.77 & \textbf{0.492} \\
            \bottomrule
        \end{tabularx}
        \begin{tablenotes}
            \item[a] $\times$ denotes that the system totally failed.
            \item[b] The bold values stand for the best result of each data sequence.
        \end{tablenotes}
    \end{threeparttable}
\end{table}

\subsection{Frame segmentation}\label{chap_82_frame_segmentation}

% \begin{figure*}
% 	\centering
% 	\includegraphics[width=\textwidth]{figs/fig_6_obervability_path.pdf}
% 	\caption{Map of point cloud observability trajectories with trajectories colored by normalized observability scores. The higher the observability score, the redder the color and the more degenerate the point cloud is, and the regions with higher observability scores are usually long, straight corridors with fewer features. By analyzing the observability, we can obtain both degenerate and non-degenerate regions in the scene.}
% 	\label{fig_6_obervability_path}
% \end{figure*}

\begin{figure}[h]
	\centering
	\includegraphics[width=\columnwidth]{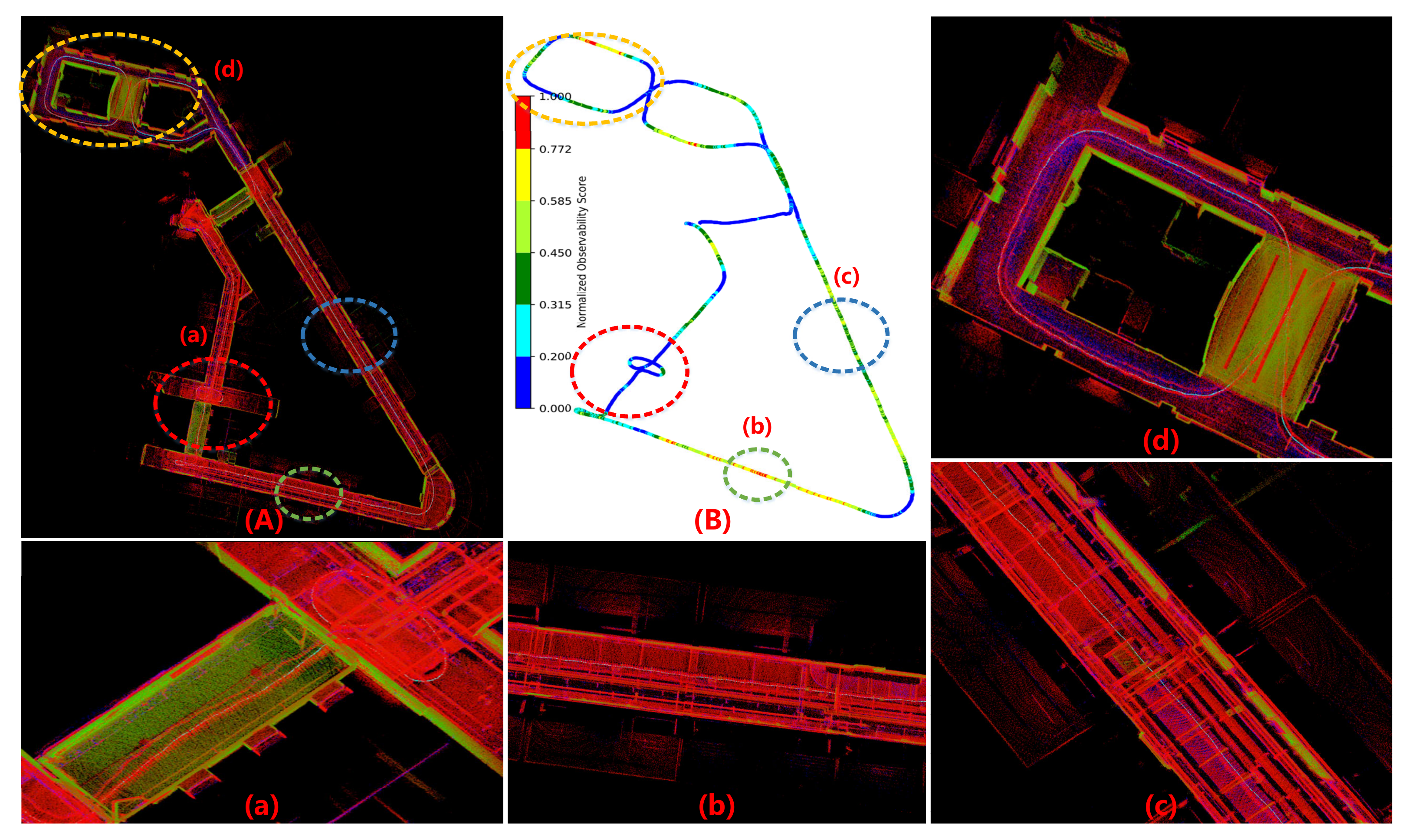}
	\caption{Map of point cloud observability trajectories with trajectories colored by normalized observability scores. The higher the observability score, the redder the color and the more degenerate the point cloud is, and the regions with higher observability scores are usually long, straight corridors with fewer features. By analyzing the observability, we can obtain both degenerate and non-degenerate regions in the scene.}
	\label{fig_6_obervability_path}
\end{figure}
To quantify changes in point cloud observability due to variations in feature richness or poverty resulting from point cloud, we draw inspiration from \cite{gelfand2003geometrically} and calculate the observability score $\alpha$ through (\ref{eq_10_factor}), which computes the ratio of the maximum eigenvalue to the minimum eigenvalue of the observability matrix $\mathbf{A}$. Additionally, we employ the CUMT-Coal-Mine dataset, which consists of three straight tunnels and multiple intersections (see Fig. \ref{fig_6_obervability_path}(B)), to evaluate our algorithm's performance in tunnel scenarios. Fig. \ref{fig_6_obervability_path} illustrates examples of normalized observability scores across the dataset. We observe that long corridors (b) and (c) exhibit higher observability scores, whereas intersections (a) and feature-rich areas (d) demonstrate lower observability scores.

\begin{figure}[h]
	\centering
	\includegraphics[width=\columnwidth]{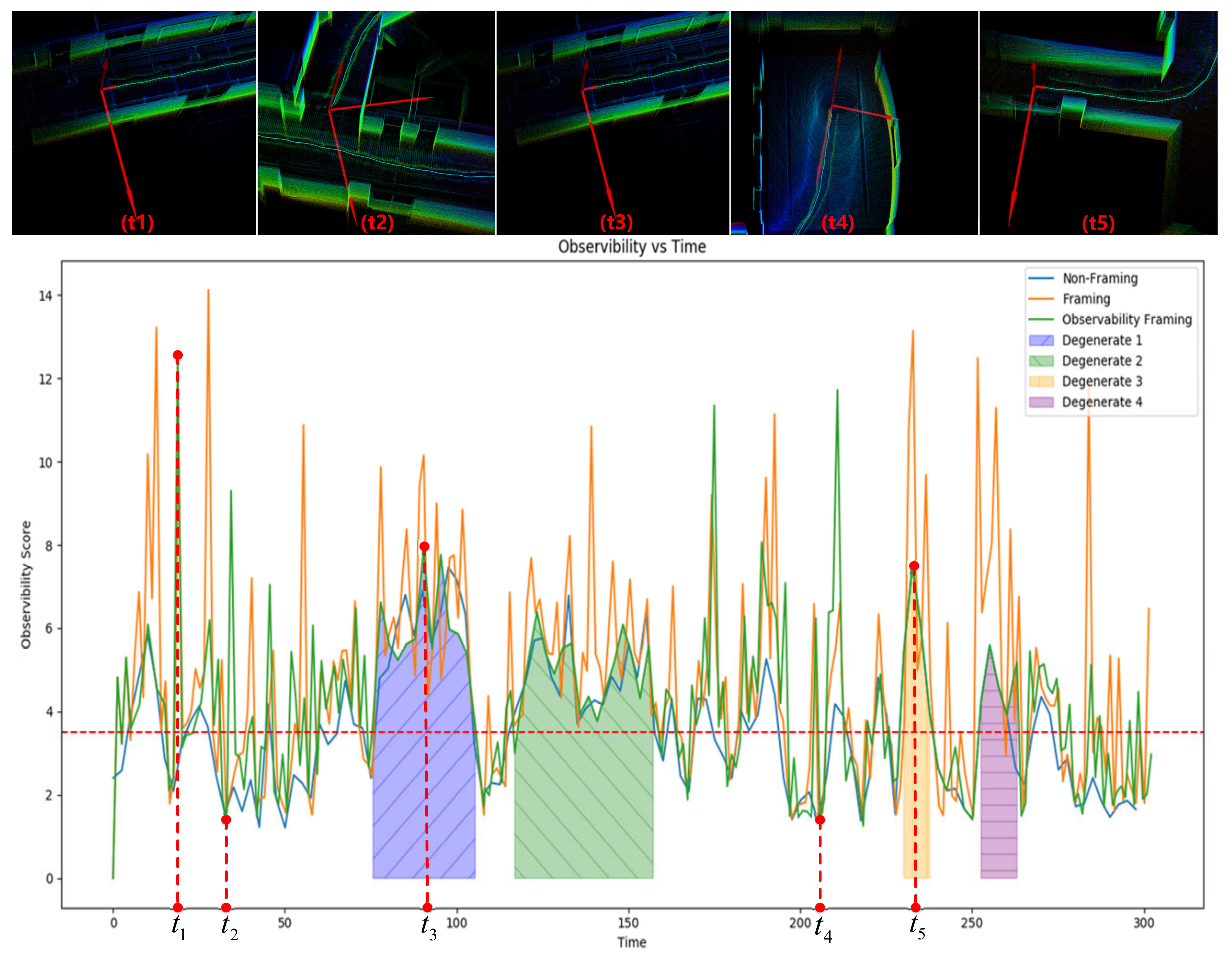}
	\caption{observability scores $\alpha$ in the CUMT dataset plots (bottom), and the\ five snapshots of the represented eigenvectors represented by $\alpha$ (scaled by the corresponding eigenvalues) (top).}
	\label{fig_7_obervability_analysis}
\end{figure}

To further quantitatively describe observability across frames, we visualize the eigenvectors of the current observability matrix of the point cloud and analyze the algorithm's performance under no segmentation, segmentation, and observability segmentation conditions. As shown in Fig. \ref{fig_7_obervability_analysis}, the blue, orange, and green curves represent the observability score results for no segmentation, segmentation, and observability segmentation, respectively. We use $\alpha=3.5$ (refer to \ref{eq_12_d_map}) as the threshold for distinguishing between segmentation and no segmentation, or degenerate and non-degenerate.

At the beginning and end of the corridor, where there are sufficient features in all directions, the observability score is approximately $\alpha \approx 2$. However, in other areas of the corridor, the observability varies with different scores. At $t = t_1$ , when the robot enters the first corridor, the system transitions from segmentation to no segmentation, resulting in an observability score of $\alpha > 10$. This high observability score makes it difficult to determine if the robot is moving based on changes in the structure of the point cloud. When $t=t_2$ the robot reaches an intersection in the corridor, and the observability improves. The system transitions from no segmentation to segmentation, with an observability score of $\alpha \approx 1.3$. At $t=t_3$ , as the robot enters the second corridor, observability weakens, and the system transitions to no segmentation, with an observability score of $\alpha \approx 8.3$. At $t=t_4$, when the robot reaches another intersection in the corridor, observability increases, and the algorithm transitions from no segmentation to segmentation, with an observability score of $\alpha \approx 1.25$. Finally, at $t=t_5$, the robot enters the third corridor, where the observability decreases. Consequently, the system switches from no segmentation state to segmentation state, with an observability score of $\alpha \approx 7.6$. It's worth noting that our score curve of adaptive observability segmentation remains consistent with the score curve of segmentation when the scene does not degenerate. However, when degeneration occurs, it aligns with the no segmentation score curve. This consistency may be why our algorithm outperforms others.

To evaluate the robustness and accuracy of our adaptive observability segmentation method across different scenarios, we conducted ablation experiments using the open-source datasets MCD \cite{thien2024mcd} and BotanicGarden \cite{liu2024botanicgarden}. The results are summarized in Table \ref{tbl2}.
% \begin{table*}[h]
% 	\centering
% 	\label{tab:tbl2}
% 	\begin{threeparttable}
% 		\caption{\textsc{ABLATION STUDY OF FRAME SEMENTATION ON RMSE OF ATE.}}\label{tbl2}
% 		%		\begin{tabular*}{\linewidth}{@{}LLLLLL@{}}
% 		\begin{tabularx}{\linewidth}{@{}l *{5}{X}@{}} %尽量保证单列，除非数据过多
% 			\toprule
% 			Dataset & Ours w/o frame segmentation & Ours frame segmentation & Ours adaptive frame segmentation & FAST-LIO2 & LIO-SAM\\
% 			\midrule
% 			MCD-tuhh07 & 2.52 & 2.43 & $^{b}$\textbf{1.53} & 2.24 & 2.56 \\
% 			MCD-tuhh08 & 6.35 & 5.34 & \textbf{3.25} & 4.32 & 5.45 \\
% 			MCD-tuhh09 & 2.42 & 2.34 & \textbf{2.33}	& 2.56 & ${}^{a}$ $\times$ \\
% 			BotanicGarden-1005-01 &  0.70 & 0.69 & 0.66 & \textbf{0.56} & 1.54 \\
% 			BotanicGarden-1018-00 & 0.29 & 0.25 & 0.28 & \textbf{0.22} & 1.43 \\
% 			BotanicGarden-1018-13 & 0.32 & 0.33 & \textbf{0.29} & 0.34 & 0.78 \\
% 			\bottomrule
% 		\end{tabularx}
% 		\begin{tablenotes}
% 			\item[a] $\times$ denotes that the system totally failed.
% 			\item[b] The bold values stand for the best result of each data sequence.
% 		\end{tablenotes}
% 	\end{threeparttable}
% \end{table*}

\begin{table*}[h]
	\centering
	\label{tab:tbl2} 
	\begin{threeparttable}
		\caption{\textsc{ABLATION STUDY OF FRAME SEMENTATION ON RMSE OF ATE.}}\label{tbl2}
		%		\begin{tabular*}{\linewidth}{@{}LLLLLL@{}}
		\begin{tabularx}{\linewidth}{@{}l *{6}{X}@{}} %尽量保证单列，除非数据过多
			\toprule
			Algorithm & MCD-tuhh07 & MCD-tuhh08 & MCD-tuhh09 & BotanicGarden-1005-01 & BotanicGarden-1018-00 & BotanicGarden-1018-13\\
			\midrule
			Ours w/o frame segmentation & 2.52 & 6.35
			& 2.42 & 0.70 & 0.29 & 0.32\\
			Ours frame segmentation & 2.43 & 5.34 & 2.34 & 0.69 & 0.25 & 0.33\\
			Ours adaptive frame segmentation & \textbf{1.53} & \textbf{3.25} & \textbf{2.33} & 0.66 & 0.28 & \textbf{0.29}\\
			FAST-LIO2 &  2.24 & 4.32 & 2.56 & \textbf{0.56} & \textbf{0.22} & 0.34 \\
			LIO-SAM & 2.56 & 5.45 & \texttimes & 1.54 & 1.43 & 0.78\\
			DLIO & 5.63 & 4.07 & 3.35 & 0.89 & 0.35 & 0.59\\
			IG-LIO & 2.01 & 3.61 & 2.83 & 135.9 & 3.15 & 10.05 \\
			Point-LIO & 3.81 & 3.71 & 3.66 & 1.32 & 0.97 & 0.97\\
			\bottomrule
		\end{tabularx}
		\begin{tablenotes}
			\item[a] $\times$ denotes that the system totally failed.
			\item[b] The bold values stand for the best result of each data sequence.
		\end{tablenotes}
	\end{threeparttable}
\end{table*}

On the four datasets "MCD-tuhh07," "MCD-tuhh08," "MCD-tuhh09," and "BotanicGarden-1018\_13" our adaptive frame segmentation method outperformed our frame segmentation, our method without frame segmentation, as well as FAST-LIO2, LIO-SAM, DLIO, IG-LIO and Point-LIO. Notably, on the "MCD-tuhh07" dataset, our method achieved an accuracy of 1.53, which was significantly better than all other methods. In the "BotanicGarden-1005\_01" and "BotanicGarden-1018-00" datasets, our adaptive frame segmentation achieved nearly the same accuracy as FAST-LIO2, and outperformed LIO-SAM, DLIO, IG-LIO, and Point-LIO.

Therefore, we conclude that our adaptive frame segmentation method maintains relatively high accuracy across various datasets, outperforming our frame segmentation, our method without frame segmentation, FAST-LIO2, LIO-SAM, DLI, IG-LIO and Point-LIO.

\subsection{Over-range detection}\label{chap_83_overange_detection}
% 大图位置进行要调整到和该章节在同一页
\begin{figure}
	\centering
	\includegraphics[width=\columnwidth]{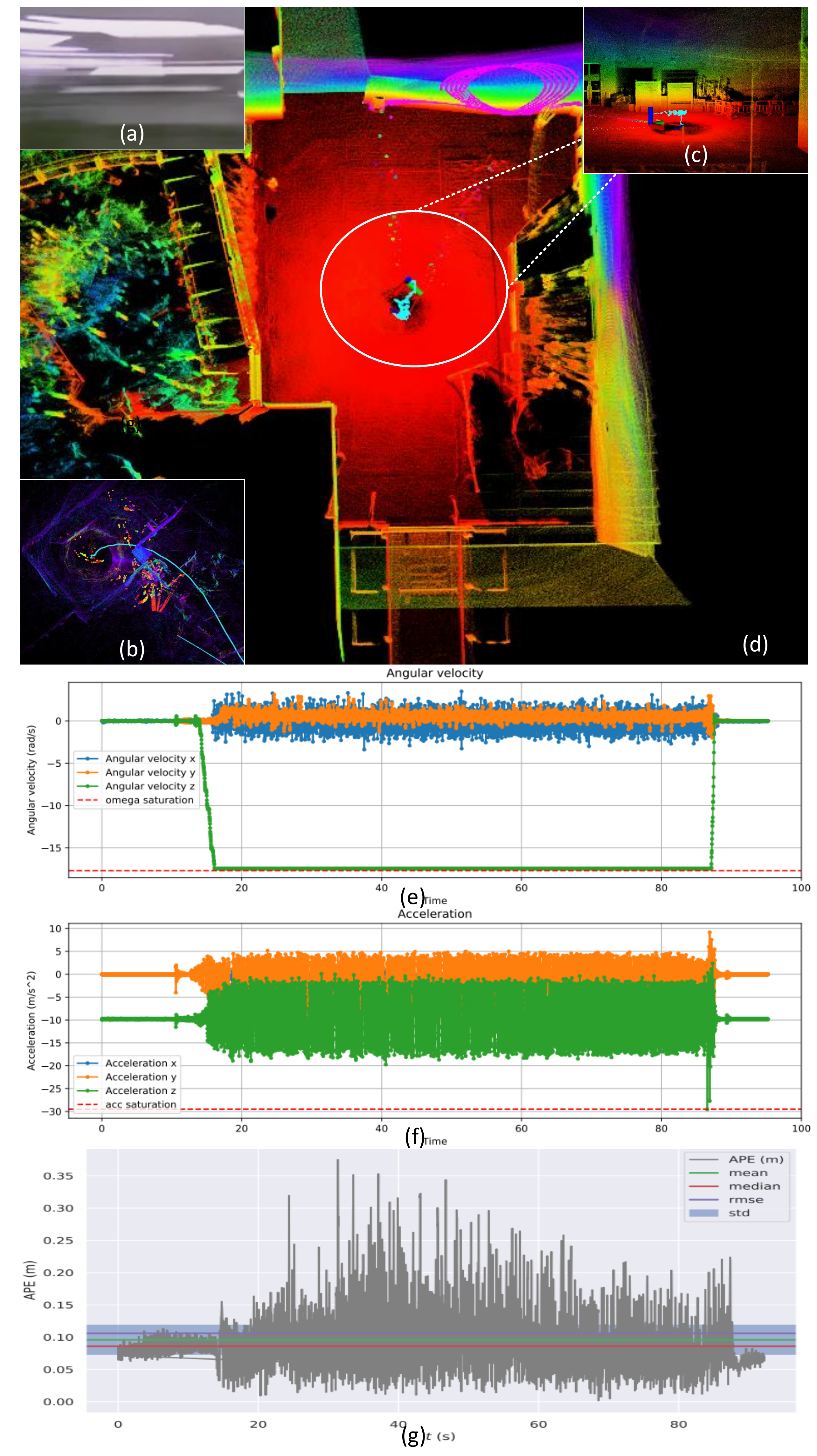}
	\caption{Spinning\_platform dataset. It shows when the acceleration exceeds the maximum IMU range (29.5 $m/s^2$), or the angular velocity reaches the maximum IMU angular velocity range (17.5 $rad/s$) and (a) the camera is completely blurred in fast motion. (b) FAST-LIO2 drifts significantly (c) while our method is still valid.}
	\label{fig_8_overange_detect}
\end{figure}

To evaluate Adaptive-LIO, we used the Point-LIO datasets spinning\_platform and PULSAR to test the performance of our method, as well as our method without over-range detection, under challenging environments. The spinning\_platform dataset has acceleration exceeding the IMU's maximum range of 3G (29.5 $m/s^2$) and angular velocity reaching the IMU's maximum range of 17.5 $rad/s$, while the PULSAR dataset has angular velocity reaching 34.8 $rad/s$. (Note: Since PULSAR lacks ground truth, we only tested the accuracy of each algorithm on the spinning\_platform dataset.)

In the spinning\_platform dataset, Fig. \ref{fig_8_overange_detect}(c),(d) show the mapping results of Adaptive-LIO, while Fig. \ref{fig_8_overange_detect}(c),(d) display the recorded IMU data over time. Fig. \ref{fig_8_overange_detect}(g) compared the accuracy of Adaptive-LIO and Point-LIO in this scenario. Notably, even during extreme motion, our method demonstrated minimal drift. This performance can be attributed to the adaptive motion mode switching in our method. From 0 to 15 seconds, the IMU was operating normally, and the system used the LIO mode for state estimation. At 16 seconds, the IMU became saturated, and using the IMU data at that point would cause the system to crash (as shown in Table 3, FAST-LIO2, Adaptive-LIO (without over-range detection), DLIO, and IG-LIO all crashed). At this moment, the system switched from LIO to LO mode for state estimation. At 89 seconds, when the IMU returned to normal, the system switched back from LO to LIO mode.  As shown in Fig. \ref{fig_8_overange_detect}(g), we compared the trajectory accuracy between our method and Point-LIO. Our method exhibited only a 0.12 RMSE difference when evaluated with APE compared to Point-LIO, demonstrating accurate state estimation in this challenging scenario.

In the PULSAR dataset, as shown in Table \ref{tbl_chap_83_overange}, only Adaptive-LIO and Point-LIO were able to run successfully, while Adaptive-LIO (w/o over-range detection) and other algorithms failed in this scenario.

\begin{table}[h]
    \centering
    \caption{COMPARISON WITH POINT-LIO DATASET.}\label{tbl_chap_83_overange}
    \begin{threeparttable}
        \begin{tabularx}{\linewidth}{@{}l *{7}{X}@{}}
            \toprule
             & \rotatebox{45}{Adaptive-LIO} & \rotatebox{45}{${}^{c}$Adaptive-LIO$^*$} & \rotatebox{45}{FAST-LIO2} & \rotatebox{45}{Point-LIO} & \rotatebox{45}{LIO-SAM} & \rotatebox{45}{DLIO} & \rotatebox{45}{IG-LIO} \\
            \midrule
            spinning\_platform & ${}^{a}$\checkmark & ${}^{b}$ $\times$ & $\times$ & \checkmark & $\times$ & $\times$ & $\times$\\
			PULSAR & \checkmark & $\times$ & $\times$ & \checkmark & $\times$ & $\times$ & $\times$\\
            \bottomrule
        \end{tabularx}
        \begin{tablenotes}
            \item[a] \checkmark denotes that the system run successfully.
            \item[b] $\times$ denotes that the system totally run failed.
            \item[c] Adaptive-LIO$^*$ denotes Adaptive-LIO (w/o over-range detection).
        \end{tablenotes}
    \end{threeparttable}
\end{table}

%\newpage
\subsection{Multi-resolution maps}\label{chap_84_multiresolution_map}

\begin{figure*}[ht]
	\centering
	\includegraphics[width=\textwidth]{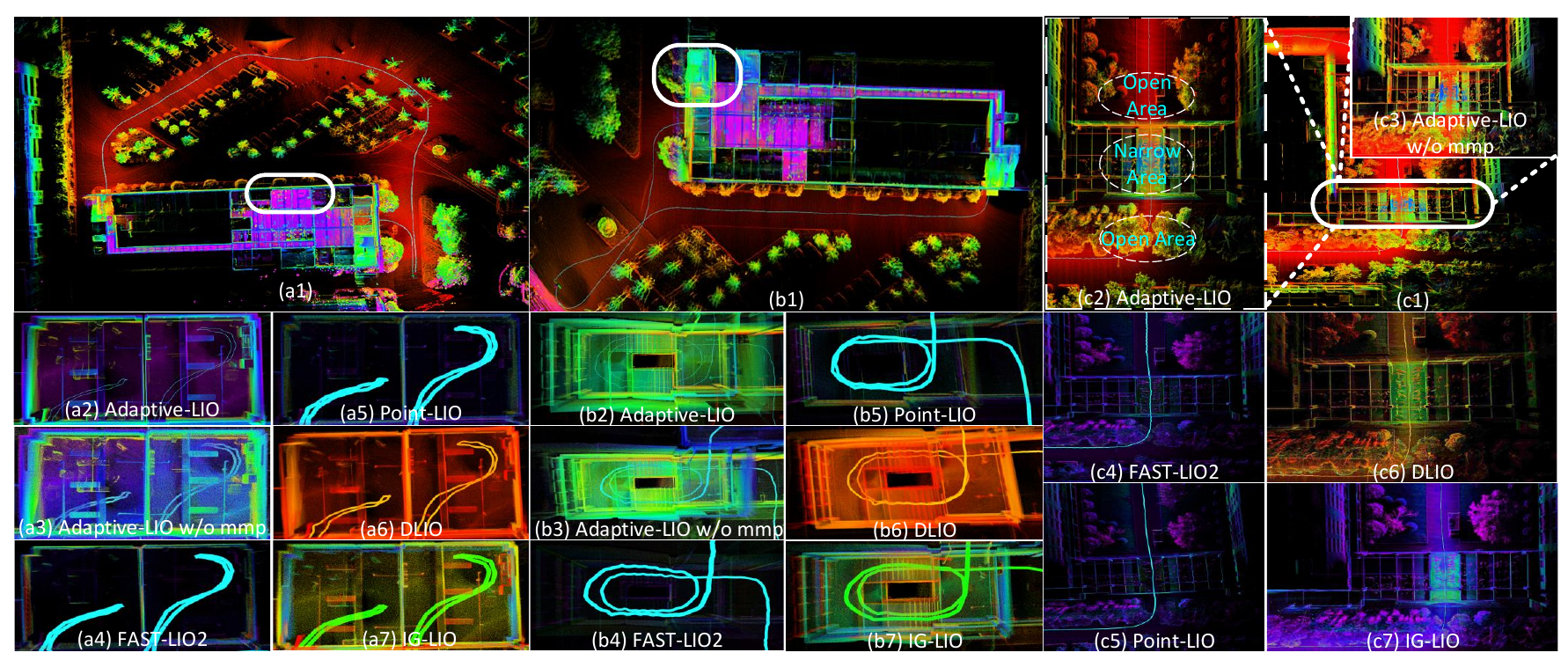}
		\caption{Robot Town. A detailed map of Robot Town generated by Adaptive-LIO, including (A) Building 1, (B) Building 2, and (C) Building 3. The mapping details of Adaptive-LIO, Adaptive-LIO w/o mmap, and other algorithms (FAST-LIO2, Point-LIO, DLIO, and IG-LIO) are shown through the aerial views (a1), (b1), (c1) and close-up views (a2)-(a7), (b2)-(b7), (c2)-(c7). Adaptive-LIO achieved the best or second-best end-to-end error across all three datasets, and its mapping quality is superior to Adaptive-LIO w/o mmap and other algorithms. (Note: \textbf{"Adaptive-LIO w/o mmap"} refers to the performance of our algorithm without the multi-resolution map.)}
		\label{fig_9_multimap_exp}
	\end{figure*}

	\begin{figure}[h]
		\centering
		\includegraphics[width=0.6\columnwidth]{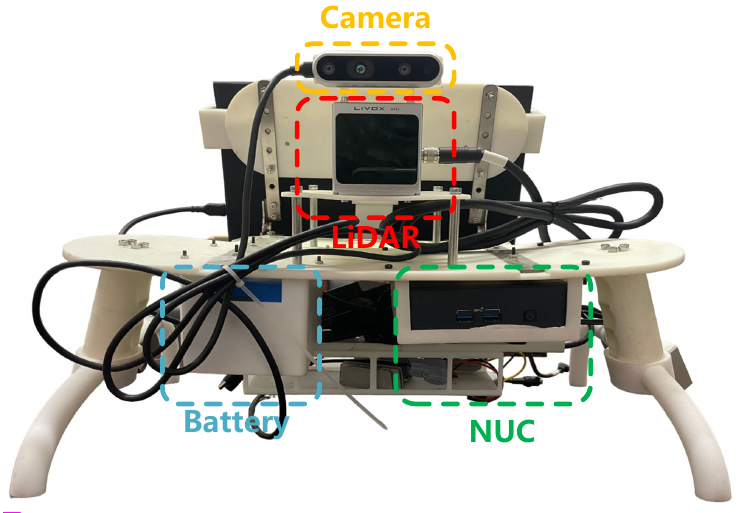}
		\caption{Our handheld device for data collection.}
		\label{fig_10_hand_device}
	\end{figure}

We collected three large-scale datasets in Robot Town-Xiaoshan District, Hangzhou, for comparison in Fig. \ref{fig_9_multimap_exp}. These datasets were collected using our handheld platform Fig. \ref{fig_10_hand_device} with a total trajectory length of 3225.6 meters. Our sensor suite includes Livox Avia and Realsense-D435i.To ensure fairness in comparison, as shown in Table \ref{tbl_chap_84_parameter}, Adaptive-LIO uses an adaptive map resolution set (0.2, 0.5, 1.2), while Adaptive-LIO (without multi-resolution map), FAST-LIO2, Point-LIO, DLIO, and IG-LIO use a fixed resolution of 0.5. Please note that due to the lack of ground truth, absolute trajectory error comparison was not possible. Therefore, as per conventional practice, we computed the end-to-end translation error as the measurement metric (Table \ref{tbl_chap_84_translation_error}). In these experiments, Adaptive-LIO excelled in both end-to-end translation error and efficiency in all aspects. The maps generated by Adaptive-LIO accurately reflect map details, providing more precise geographic information for autonomous mobile robots.

\begin{table}[h]
    \centering
    \caption{PARAMETER SETTINGS of OUR PROPOSED APDAPTIVE MULTI-RESOLUTIONMAP SETTING STRATEGY (UNIT: M).}\label{tbl_chap_84_parameter}
    \begin{threeparttable}
        \begin{tabularx}{\linewidth}{@{}l *{7}{X}@{}}
            \toprule
              & \rotatebox{45}{Adaptive-LIO} & \rotatebox{45}{${}^{a}$Adaptive-LIO${}^{+}$} & \rotatebox{45}{FAST-LIO2} & \rotatebox{45}{Point-LIO} & \rotatebox{45}{LIO-SAM} & \rotatebox{45}{DLIO} & \rotatebox{45}{IG-LIO} \\
            \midrule
            Resolution & (0.2, 0.5, 1.2) & 0.5 & 0.5 & 0.5 & 0.5 & 0.5 & 0.5\\
            \bottomrule
        \end{tabularx}
        \begin{tablenotes}
            \item[a] Adaptive-LIO${}^{+}$ denotes Adaptive-LIO (w/o multi-resolutionmap).
        \end{tablenotes}
    \end{threeparttable}
\end{table}

% ===== 精度 + 耗时 ====
\begin{table}[h]
	\centering
	\caption{\textsc{End-to-End Translation Error Comparison with Robot Town Dataset (UNIT:M).}}
	\label{tbl_chap_84_translation_error}
	\small % 减小字体大小以适应更多内容
	\setlength\tabcolsep{3pt} % 减少列间隔
	\begin{tabular*}{\linewidth}{@{}lp{1.8cm}p{1.8cm}p{1.8cm}@{}}
		\toprule
		\multirow{2}{*}{\shortstack[c]{}} & \multicolumn{3}{c}{\shortstack[c]{End-to-End\\Translation Error}} \\
		\cmidrule(lr){2-4} % 控制分割线范围
		& {A} & {B} & {C} \\
		\midrule
		Adaptive-LIO & \textbf{1.21} & \textbf{1.36} & 1.38 \\
		Adaptive-LIO(w/o mmap) & 1.43 & 1.76 & 1.51 \\
		FAST-LIO2 & 6.45 & 16.9 & 1.94 \\
		Point-LIO & 2.38 & 3.83 & \textbf{1.37} \\
		LIO-SAM & $\times$ & $\times$ & $\times$ \\
		DLIO & 12.38 & 1.40 & 1.66 \\
		IG-LIO & 1.32 & 1.39 & $\times$ \\
		\bottomrule
	\end{tabular*}
\end{table}

\begin{table}[h]
	\centering
	\caption{\textsc{TIME CONSUMPTION PER FRMAE with Robot Town Dataset (UNIT: MS).}}
	\label{tbl_chap_84_computation_time}
	\small % 减小字体大小以适应更多内容
	\setlength\tabcolsep{3pt} % 减少列间隔
	\begin{tabular*}{\linewidth}{@{}lM{1.8cm}M{1.8cm}M{1.8cm}@{}}
		\toprule
		\multirow{2}{*}{\shortstack[c]{}} & \multicolumn{3}{c}{\shortstack[c]{Average\\Computation Time}} \\
		\cmidrule(lr){2-4}
		& {A} & {B} & {C} \\
		\midrule
		Adaptive-LIO & 33.59 & 31.47 & 28.56 \\
		Adaptive-LIO(w/o mmap) & 27.67 & 26.21
		& 23.57\\
		FAST-LIO2 & 9.36 & 10.49 & 11.33\\
		Point-LIO & 7.98 & 11.31 & 10.54\\
		LIO-SAM & 55.76 & 57.18 & 59.45\\
		DLIO & 5.92 & 6.32 & 5.34\\
		IG-LIO & \textbf{4.75} & \textbf{5.28} & \textbf{4.69}\\
		\bottomrule
	\end{tabular*}
\end{table}

In experiments (A) and (B), we evaluated the performance of Adaptive-LIO during the transition from outdoor to indoor environments. Fig. \ref{fig_9_multimap_exp}(a2) and (b2) show the performance of Adaptive-LIO in narrow rooms and hallways, while Fig. \ref{fig_9_multimap_exp}(a3) and (b3) display the performance of Adaptive-LIO without multi-resolution map management (w/o mmap) in the same environments. Fig. \ref{fig_9_multimap_exp}(a4)-(a7) and  Fig. \ref{fig_9_multimap_exp}(b4)-(b7) provide detailed comparisons with FAST-LIO2, Point-LIO, DLIO, and IG-LIO. In these narrow areas, Adaptive-LIO's mapping accuracy outperformed Adaptive-LIO w/o mmap, FAST-LIO2, Point-LIO, DLIO, and IG-LIO, capturing finer details of the environment's texture more effectively.

In experiment (C), we tested Adaptive-LIO in a large outdoor scene. The dashed lines indicate a zoomed-in area where the environment transitions from an open area to a narrow one. Fig. \ref{fig_9_multimap_exp}(c2)-(c6) show the detailed results of Adaptive-LIO, Adaptive-LIO w/o mmap, FAST-LIO2, Point-LIO, DLIO, and IG-LIO in this region. During the transition from an open area to a narrow space, the voxel resolution shifts from high to low, allowing Adaptive-LIO to produce clearer point clouds. The mapping results of Adaptive-LIO w/o mmap and other algorithms fail to clearly capture the features of the narrow area. When transitioning from narrow to open areas, where the voxel resolution switches from low to high, Adaptive-LIO continues to demonstrate superior point cloud accuracy in open regions. LIO-SAM failed in all experiments (A), (B), and (C).

We also evaluated the average processing time per frame (in milliseconds) for the system across all experiments. As shown in Table \ref{tbl_chap_84_computation_time}, Adaptive-LIO took 25-35 ms per frame in experiments (A), (B), and (C), which was longer than the processing times of Adaptive-LIO w/o multi-resolution, FAST-LIO2, Point-LIO, DLIO, and IG-LIO. This is because Adaptive-LIO manages multi-resolution map, involving frequent insertions and deletions, making it less efficient than Kalman filter-based methods. Although other algorithms achieved better time efficiency, the proposed method still meets the real-time requirements across all experiments.

% % ===== 精度 + 耗时 ====

\section{Conclusion}\label{chap_9_conclusion}
In this paper, we propose Adaptive-LIO, an observability-aware algorithm for localization and mapping that accurately and efficiently estimates odometry in GPS-restricted and unknown underground environments. The algorithm leverages adaptive segmentation to enhance mapping accuracy, adapts motion modality through IMU saturation and fault detection, and adjusts map resolution adaptively using multi-resolution voxel maps to accommodate scenes at varying distances from the LiDAR center. The developed system was thoroughly tested in real-world scenarios and open-source datasets, including extreme motion, indoor-outdoor transitions, and diverse LiDAR types, environments, and motion patterns from public datasets. Across all tests, Adaptive-LIO achieved computational efficiency and odometry accuracy comparable to other state-of-the-art LIO algorithms.

\bibliographystyle{IEEEtranN} %针对natbib来进行设计
\bibliography{IEEEabrv,cas-refs-adalio}

%bib
\begin{IEEEbiography}[{\includegraphics[width=1in,height=1.25in,clip,keepaspectratio]{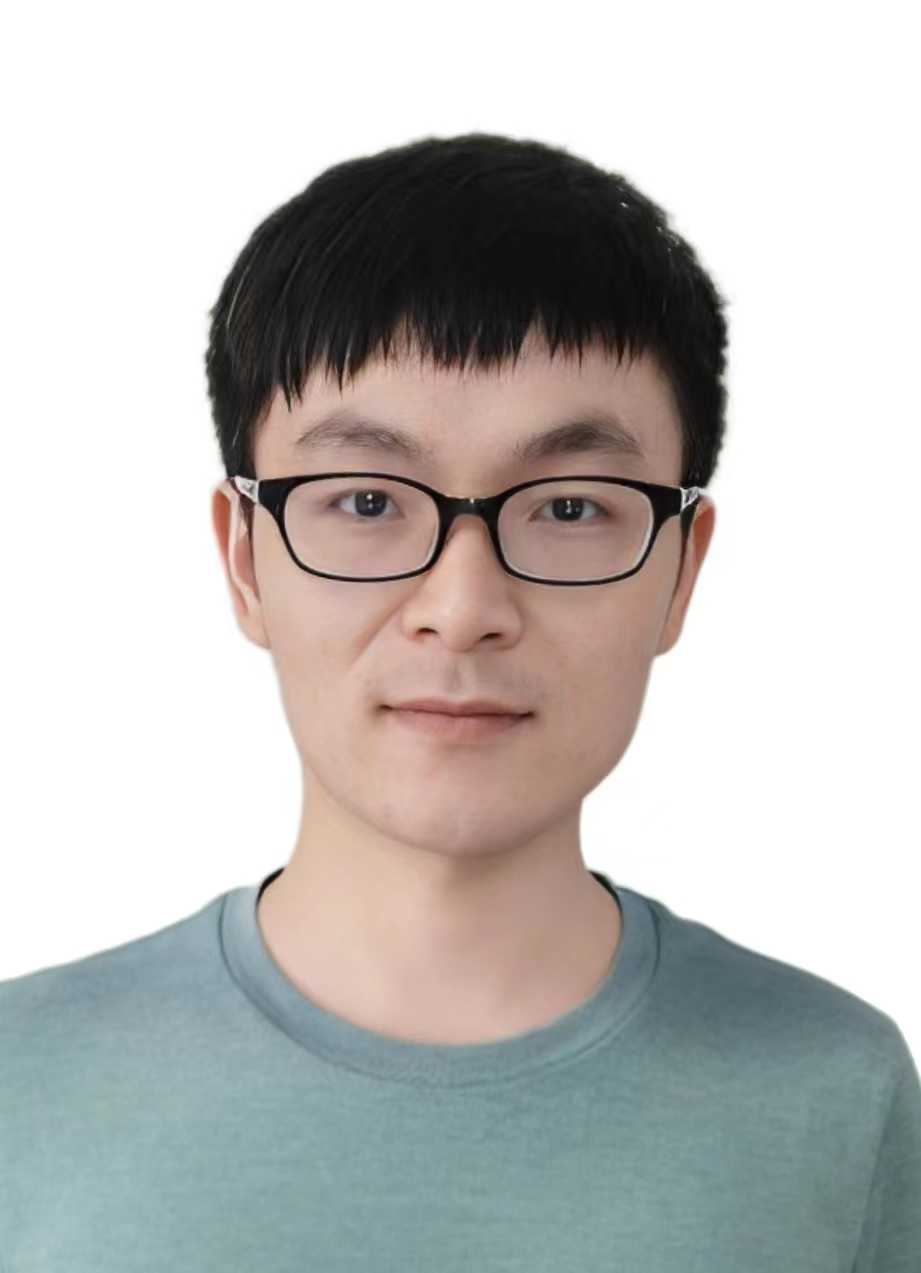}}]{Chengwei Zhao} received the B.E. from Zhoukou Normal University, Zhoukou, China in 2013 and the  M.S. degree from Jimei University in 2016. He is currently with Hangzhou Qisheng Intelligent Techology Co. Ltd. His research interests include sensor fusion, SLAM in industrial application and manifold math.
\end{IEEEbiography}

\begin{IEEEbiography}[{\includegraphics[width=1in,height=1.25in,clip,keepaspectratio]{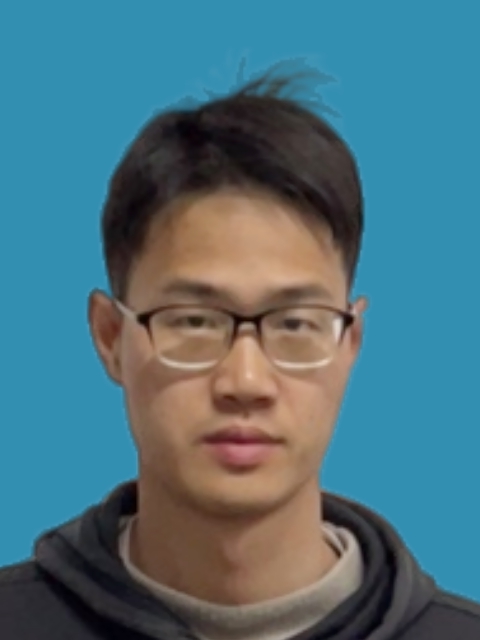}}]{Kun Hu} received the B.E. degree from XiangTan University in 2021. He is currently pursuing a M.S. degree with China University of Mining and Technology. His research interests include multi-sensor fusion, path planning and coal mine robot motion control.
\end{IEEEbiography}

\begin{IEEEbiography}[{\includegraphics[width=1in,height=1.25in,clip,keepaspectratio]{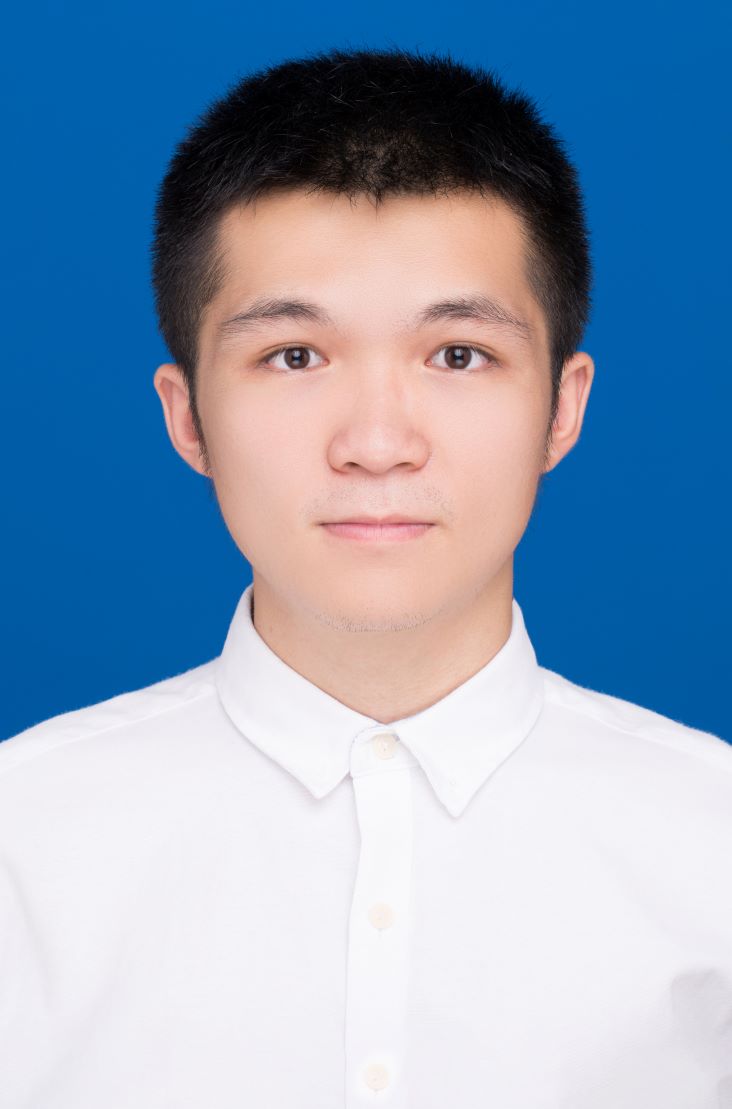}}]{Jie Xu} received the B.E., M.S. degree from the Harbin Institute of Technology in 2018, 2020 respectively. He is currently pursuing a Ph.D. degree with the School of Mechanical and Electrical Engineering, Harbin Institute of Technology. Meanwhile, he is participating in a joint doctoral training program at NTU EEE College. His research interests include multi-sensor calibration, RGB-D SLAM and multimodal SLAM.
\end{IEEEbiography}

\begin{IEEEbiography}[{\includegraphics[width=1in,height=1.25in,clip,keepaspectratio]{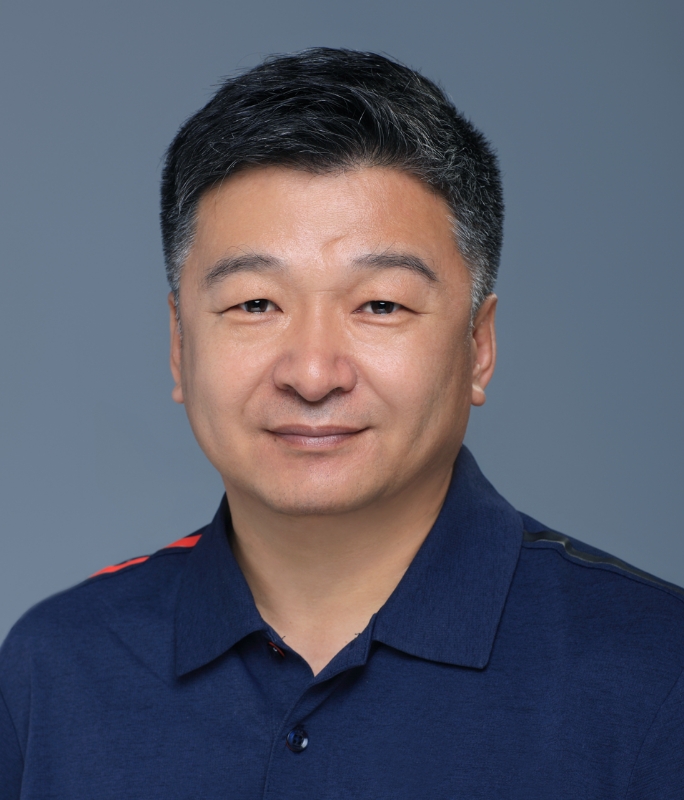}}]{Lijun Zhao} received Ph.D degree in mechatronic engineering from institute of robotics at Harbin Institute of Technology, in 2009. He is currently a Professor with the state key laboratory of robotics and system.
\end{IEEEbiography}

\begin{IEEEbiography}[{\includegraphics[width=1in,height=1.25in,clip,keepaspectratio]{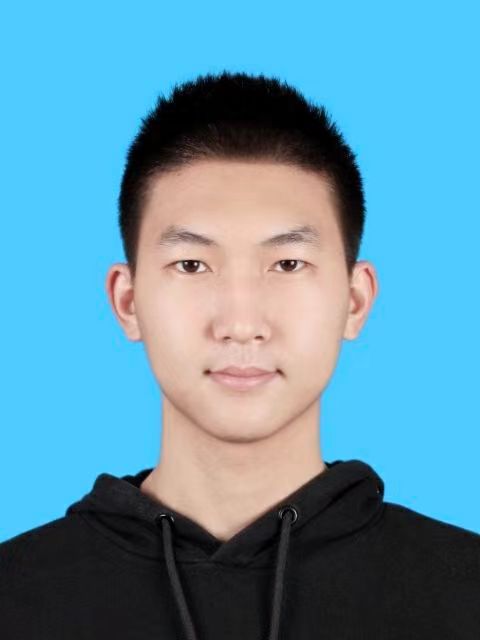}}]{Baiwen Han} was born in Liaoning, China in 2001. He received his B.E. degree from the central south university, Changsha, China in 2023. He is currently pursuing an M.S. degree with the State Key Laboratory of Robotics and Systems. His research interests include Lidar SLAM and multisensor fusion.
\end{IEEEbiography}

\begin{IEEEbiography}[{\includegraphics[width=1in,height=1.25in,clip,keepaspectratio]{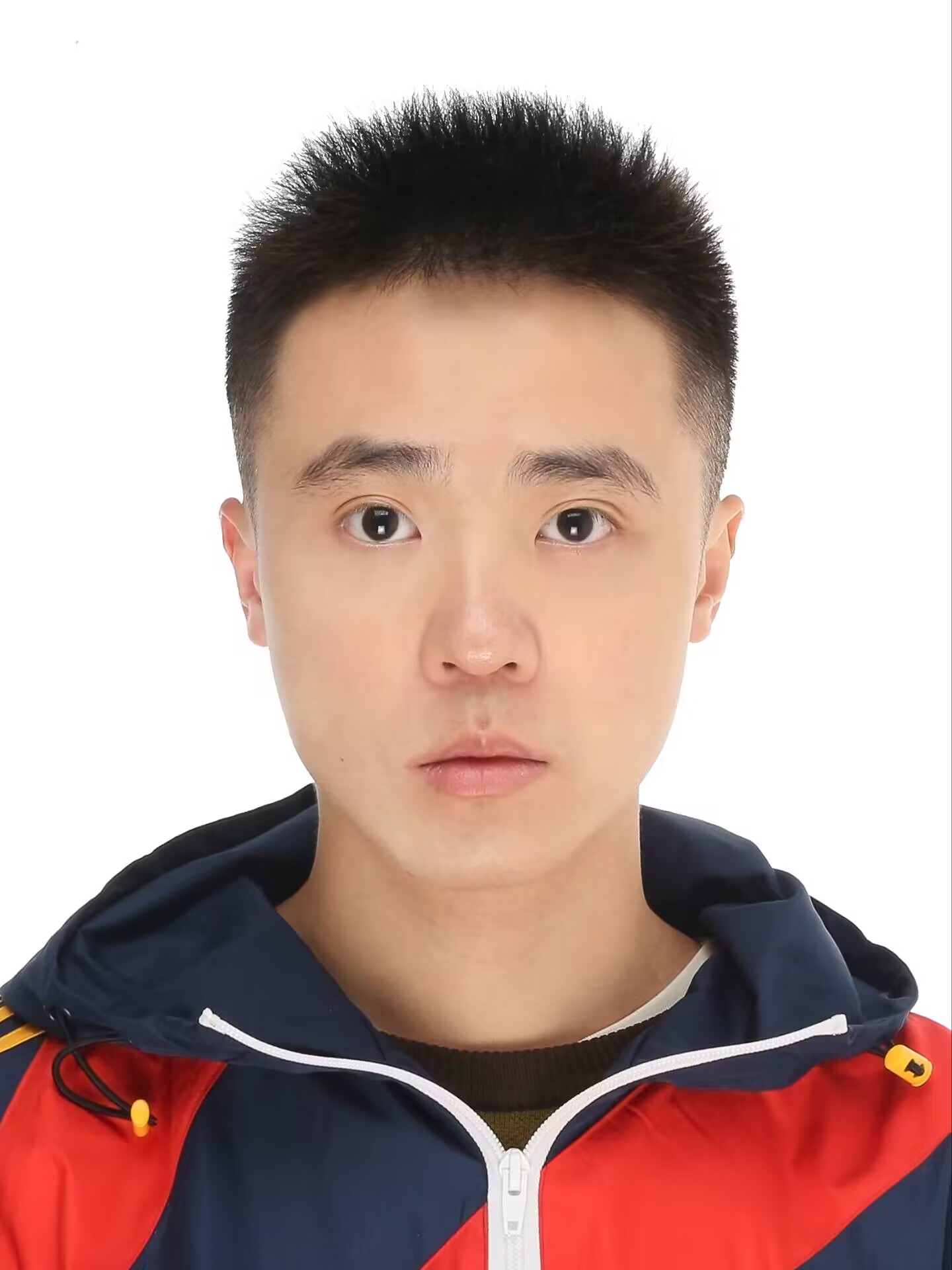}}]{Maoshan Tian} was born in Leshan, Sichuan in 1999. He is currently pursuing the M.S. degree in School of Mechanical and Electrical Engineering at the University of Electronic Science and Technology of China. His research direction is Lidar SLAM and multisensor fusion SLAM technology.
\end{IEEEbiography}

\begin{IEEEbiography}[{\includegraphics[width=1in,height=1.25in,clip,keepaspectratio]{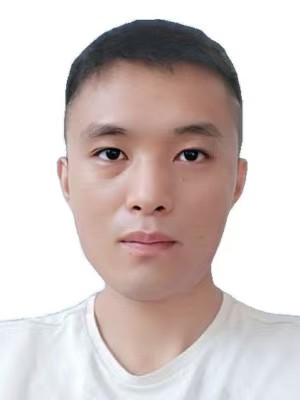}}]{Kaidi Wu} was born in Jiangsu, China in 2000. He received the B.E. degree from Huaiyin Institute of Technology in 2023 and is currently pursuing the M.S. degree in the School of Electromechanical Engineering at China University of Mining and Technology. His research interests include LiDAR SLAM and deep learning.
\end{IEEEbiography}

\begin{IEEEbiography}[{\includegraphics[width=1in,height=1.25in,clip,keepaspectratio]{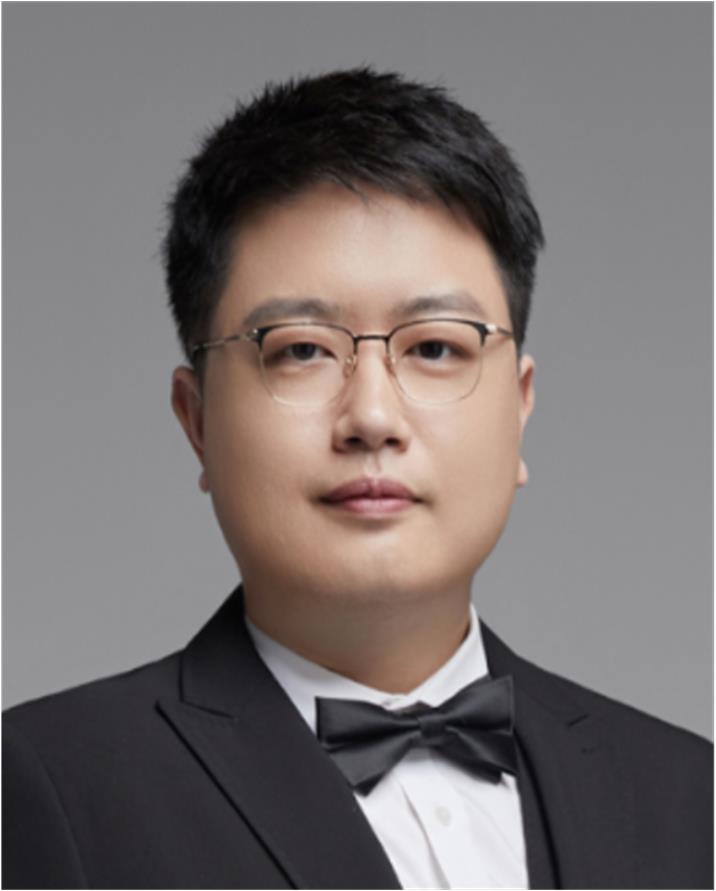}}]{Shenghai Yuan} received his B.S. and Ph.D. degrees in Electrical and Electronic Engineering in 2013 and 2019, respectively, Nanyang Technological University, Singapore. His research focuses on robotics perception and navigation. He is a postdoctoral senior research fellow at the Centre for Advanced Robotics Technology Innovation (CARTIN), Nanyang Technological University, Singapore. He has contributed over 60 papers to journals such as TRO, IJRR, TIE, RAL, and conferences including ICRA, CVPR, ICCV, NeurIPS, and IROS. Currently, he serves as an associate editor for the Unmanned Systems Journal and as a guest editor of the Electronics Special Issue on Advanced Technologies of Navigation for Intelligent Vehicles. He achieved second place in the academic track of the 2021 Hilti SLAM Challenge, third place in the visual-inertial track of the 2023 ICCV SLAM Challenge, and won the IROS 2023 Best Entertainment and Amusement Paper Award. He also received the Outstanding Reviewer Award at ICRA 2024. He served as the organizer of the CARIC UAV Swarm Challenge and Workshop at the 2023 CDC. Currently, he is the organizer of the UG2 Anti-drone Challenge and workshop at CVPR 2024. He is the organizer of the second CARIC UAV Swarm Challenge and workshop at IROS 2024.
\end{IEEEbiography}

\end{document}